\documentclass[preprint,12pt]{elsarticle}

\usepackage{lineno,hyperref}
\usepackage{natbib}
\usepackage{geometry}
\usepackage{fleqn}
\usepackage{graphicx}
\usepackage{newtxtext,newtxmath}
\usepackage{hyperref}
\usepackage{booktabs}
\usepackage{bm}
\usepackage{amsmath}
\usepackage{enumitem}
\usepackage{esvect}
\usepackage{soul,xcolor}
\usepackage{caption}
\usepackage{subcaption}
\usepackage[T1]{fontenc}
\modulolinenumbers[5]
\usepackage{comment}

\graphicspath{{./figures/}}
\journal{Elsevier}

\begin{document}

\begin{frontmatter}

\title{\large{Towards Overcoming Data Scarcity in Nuclear Energy: A Study on Critical Heat Flux with Physics-consistent Conditional Diffusion Model}}

\author[NCSU,PSI]{Farah Alsafadi}
\author[NCSU]{Alexandra Akins}

\author[NCSU]{Xu Wu\corref{mycorrespondingauthor}}
\cortext[mycorrespondingauthor]{Corresponding author}
\ead{xwu27@ncsu.edu}

\address[NCSU]{Department of Nuclear Engineering, North Carolina State University    \\
Burlington Engineering Laboratories, 2500 Stinson Drive, Raleigh, NC 27695, USA \\}
\address[PSI]{Laboratory for Reactor Physics and Thermal-Hydraulics, \\
Center for Nuclear Engineering and Sciences, Paul Scherrer Institute, 5232 Villigen PSI, Switzerland}

\begin{abstract}

Deep generative modeling provides a powerful pathway to overcome data scarcity in energy-related applications where experimental data are often limited, costly, or difficult to obtain. By learning the underlying probability distribution of the training dataset, deep generative models, such as the diffusion model (DM), can generate high-fidelity synthetic samples that statistically resemble the training data. Such synthetic data generation can significantly enrich the size and diversity of the available training data, and more importantly, improve the robustness of downstream machine learning models in predictive tasks.
The objective of this paper is to investigate the effectiveness of DM for overcoming data scarcity in nuclear energy applications. By leveraging a public dataset on critical heat flux (CHF) that cover a wide range of commercial nuclear reactor operational conditions, we developed a DM that can generate an arbitrary amount of synthetic samples for augmenting of the CHF dataset. Since a vanilla DM can only generate samples randomly, we also developed a conditional DM capable of generating targeted CHF data under user-specified thermal-hydraulic conditions. The performance of the DM was evaluated based on their ability to capture  empirical feature distributions and pair-wise correlations, as well as to maintain physical consistency. 
The results showed that both the DM and conditional DM can successfully generate realistic and physics-consistent CHF data. Furthermore, uncertainty quantification was performed to establish confidence in the generated data. The results demonstrated that the conditional DM is highly effective in augmenting CHF data while maintaining acceptable levels of uncertainty.

\end{abstract}


\end{frontmatter}

%\linenumbers

\newpage

%%%%%%%%%%%%%%%%%%%%%%%%%%%%%%%%%%%%%%%%%%%%%%%%%%%%%%%%%%%%%%%%%%%%%%%%%%%%%%%%
%%%%%%%%%%%%%%%%%%%%%%%%%%%%%%%%%%%%%%%%%%%%%%%%%%%%%%%%%%%%%%%%%%%%%%%%%%%%%%%%
\section{Introduction}

Deep generative modeling \cite{salakhutdinov2015learning, ruthotto2021introduction} provides a powerful pathway to overcome data scarcity in energy-related applications where experimental data are often limited, costly, or difficult to obtain. By learning the underlying probabilistic distributions of the training dataset, deep generative models (DGMs), such as generative adversarial networks (GANs), variational autoencoders (VAEs), and diffusion model (DM), can generate high-fidelity synthetic datasets that statistically resemble the training data, thereby augmenting scarce experimental datasets. Such synthetic data generation can significantly enrich the size and diversity of the available training data, and more importantly, enhance the robustness of downstream machine learning (ML) models for predictive tasks.

The field of generative modeling was dominated by GANs between 2014 and the early 2020s. The era of GAN dominance began immediately after their introduction in a 2014 paper by Ian Goodfellow and his colleagues, which was later published in ``Communications of the ACM'' in 2020 \cite{goodfellow2020generative}.
%GANs consist of two neural network models, the generator that creates new data instances, and the discriminator that tries to distinguish whether a sample is real or synthetic. In the training of GANs the two models compete against each other to improve their performance, where the generator generates samples to deceive the discriminator, while the discriminator strives to accurately classify the data. The adversarial training continues until the generator produces data so realistic that the discriminator can no longer distinguish between real and fake samples. 
GANs have been applied for generative modeling in a wide range of disciplines, including energy-related domains. For example, Zhang and co-authors developed SolarGAN \cite{zhang2023solargan}, which can generate synthetic annual solar irradiance time series dataset on urban building facades using fisheye images. Rizzato et al. \cite{rizzato2022stress} used GANs for load scenario generation for stress testing of electrical grids. Carreon et al. \cite{carreon2023generative} employed GANs to generate realistic flame images resembling those from a combustor experiment. It was demonstrated that GANs were able to capture the statistical structure of the experimental data for a variety of flame configurations. Menon and Ranganathan \cite{menon2022generative} discussed the state-of-the-art applications of popular DGMs, including GANs, in the domains of biomaterials and energy materials. In a recent work, Nabila et al. \cite{nabila2025data} used a variant of GAN, the Wasserstein GAN for energy demand forecasting. 

In the early 2020s, the undisputed dominance of GANs began to wane with the rise of other highly effective DGMs, especially DMs. DMs are inspired by non-equilibrium thermodynamics \cite{sohl2015deep}, and they use a Markov chain of diffusion steps to slowly add random noise to the data and then learn to reverse the diffusion process to construct desired data samples from the noise. A series of works have shown that DMs beat GANs with image generation quality and training stability, in tasks including image synthesis \cite{dhariwal2021diffusion}, topology optimization \cite{maze2023diffusion}, talking-face generation \cite{stypulkowski2024diffused}, and image classification \cite{mukhopadhyay2023diffusion}. Even though DMs tend to be slower than GANs due to the need for a large number of denoising steps during sample generation in the reverse diffusion process \cite{croitoru2023diffusion}, some of the state-of-the-art text-to-image generation tools have been developed based on DMs, such as Dall-E \cite{ramesh2022hierarchical}, stable diffusion \cite{rombach2022high}, Imagen \cite{saharia2022photorealistic} and Midjourney. Additionally, DMs are adaptable to tasks beyond simple image generation and have been utilized for anomaly detection \cite{wyatt2022anoddpm} and time series imputation and forecasting \cite{tashiro2021csdi}. DMs have proven to be application-agnostic and have been applied for diverse uses such as in the medical field \cite{jiang2024fast, zhou2024super}, fault diagnosis in mechanical systems \cite{zhang2024interpretable}, and nuclear physics \cite{devlin2024diffusion}.

Most applications of DMs have focused on text, image, or video data, and their applications for data augmentation of scientific datasets, particularly in the energy sector, have been relatively limited. For example, Zhang et al. \cite{zhang2024generating} developed a physics-informed DM for generating synthetic net load data, addressing the challenges of data scarcity and privacy concerns. Jiang and co-authors \cite{jiang2025diffusion} developed a multivariable DM for transportation energy demand prediction that can capture time-series tendencies while considering contextual information (e.g., COVID-19). Some other examples can be found in inverse design of specialized materials for energy applications. Bastek and Kochmann \cite{bastek2023inverse} proved that video denoising DM trained on full-field data of periodic stochastic cellular structures can be used for inverse design of nonlinear mechanical metamaterials. Park et al. \cite{park2024inverse} developed a DM for the generation of novel crystalline porous materials.

The study of deep generative learning in nuclear energy is relatively recent and limited. In a previous work \cite{alsafadi2023deep}, we compared the performances of GANs, normalizing flows, VAEs and conditional VAEs to generate synthetic void fraction data using a classical benchmark on boiling water reactor bundle test. It was found that conditional VAEs achieved the smallest errors in generating scientific data. In a follow-up study \cite{alsafadi2025investigation}, we demonstrated that ML models trained using the augmented dataset can achieve much better predictive accuracy and reduced predictive uncertainties. In another recent study \cite{alsafadi2025predicting}, we investigated the conditional VAE model for predicting critical heat flux (CHF) with uncertainty quantification (UQ) and domain generalization. The numerical results proved that a conditional VAE performs better than conventional deep neural networks in predicting CHF and exhibits better uncertainty behavior. A similar problem was studied in \cite{nabila2025data} using conditional GANs for constrained synthetic data generation. Very recently, Liu et al. \cite{liu2025unsupervised} employed a DM for the development of an unsupervised anomaly detection method for nuclear power plants, and demonstrated that the DM outperformed autoencoders, VAEs, and GANs in detection accuracy. In the work by Barra et al. \cite{barra2025inverse}, the authors presented a workflow for the inverse design of molten salts with targeted density values. The authors used a dataset of critically evaluated molten salt densities to train a VAE coupled with a predictive deep neural network, which then can be used to generate new molten salt compositions with desired density values. This work can efficiently exploit molten salts' customizability and unlocking their advantages in energy production and energy storage.

The main objective of this work is to investigate the effectiveness of DM in generating physics-consistent data for CHF, which is one of the most important safety-related limiting quantities in nuclear energy. The CHF experimental dataset used for training is the public dataset that was used to develop the widely known 2006 Groeneveld CHF lookup table \cite{groeneveld20072006}. The dataset was curated and published by the U. S. Nuclear Regulatory Commission (NRC) \cite{groeneveld2019critical} and is the largest known CHF dataset publicly available with measurements in vertical water-cooled uniformly heated tubes. The dataset consists of 24,579 samples, and it is well suited for data augmentation studies, as it is sufficiently large to train a DGM, but still significantly smaller than the typical sample sizes for image generation tasks.

Another objective of this work is to develop DGMs for targeted data generation. One major limitation of the vanilla DM is that it can only produce synthetic samples randomly. However, in practical applications it is desirable to have the synthetic data generated at specific conditions, in this work, CHF values at certain thermal-hydraulic (TH) conditions. To this end, we have also developed a conditional DM (CDM) model that can generate CHF samples at given TH conditions. Conditional DGMs have been studied for different algorithms. For example, conditional GANs \cite{mirza2014conditional} were developed shortly after GANs. It has been used in \cite{baasch2021conditional} to generate building energy consumption data, which can be used for numerous downstream applications such as retrofit analysis, smart grid integration and optimization, and load forecasting. In another work \cite{walter2024probabilistic}, a conditional GAN was used for producing short-term time series electricity price scenarios. Similarly, conditional VAEs models have been proven to be very effective in our preliminary work on CHF \cite{alsafadi2025predicting}. Other generative learning using conditional VAEs in the energy sector including \cite{bregere2020simulating} for simulating tariff impact in electrical energy consumption profiles, \cite{zheng2023conditional} for wind power curve modeling, and \cite{li2024sensing} for photovoltaic power generation to study sensing anomalies. The development and applications of conditional DMs are relatively limited compared to conditional GANs and VAEs, simply due to the fact that DMs have only become popular since early 2020s. A few energy-related studies are still available in the literature. For example, \cite{dong2023short} for short-term wind power scenario generation, \cite{fu2024creating} for synthetic energy meter data using power meters from various buildings and countries, and \cite{wang2024customized} for customized load profiles synthesis for electricity customers. To the authors' knowledge, there has been no prior work on DM and conditional DM in the nuclear energy domain.

The performance of DM and CDM were evaluated based on how similar the synthetic samples are to the real CHF samples in the training dataset. We compared the marginal distributions of both the TH features and the CHF output, as well as the pair-wise correlations between them. Various error metrics were also applied to assess the performance of the DGMs. The results showed that both the DM and CDM models can successfully generate CHF data by accurately learning the empirical marginal distributions and correlations between features and CHF output. To establish confidence in the generated samples, we also performed a UQ analysis by leveraging the inherent variability in the CDM generation process. During generation, the CDM creates new samples from random noise, which is removed step by step according to the given conditions. By varying the initial noise input each time, the model produces different outputs for the same set of conditions, allowing us to quantify the uncertainties in the outputs. The CDM successfully generated CHF data with high accuracy under user-specified conditions.

The novelty and contribution of this work can be summarized as: (1) development of a DM for generative learning and demonstration for a critical safety-related limiting quantity in nuclear energy, (2) improvement of the DM to a CDM for targeted synthetic data generation, (3) a comprehensive and systematic performance evaluation of the DGM results using various quantitative metrics, including physics-consistence assessment, and (4) UQ analysis of the generative process to produce samples with quantified uncertainty.

The remainder of this paper is organized as follows: 
Section \ref{sec:methods} presents an overview the DM and CDM methodologies. Section \ref{sec:problem-definition} includes the problem definition of this study and describes how DM and CDM were trained and used for generative analysis. Section \ref{sec:results-DM} discusses the results obtained using the DM, while Section \ref{sec:results-CDM} presents the results for CDM, along with UQ analysis and physical-consistency evaluation. Finally, Section \ref{sec:conclusions} summarizes the findings and concludes the paper.

%%%%%%%%%%%%%%%%%%%%%%%%%%%%%%%%%%%%%%%%%%%%%%%%%%%%%%%%%%%%%%%%%%%%%%%%%%%%%%%%
\section{Methodologies}
\label{sec:methods}
%%%%%%%%%%%%%%%%%%%%%%%%%%%%%%%%%%%%%%%%%%%%%%%%%%%%%%%%%%%%%%%%%%%%%%%%%%%%%%%%

%%%%%%%%%%%%%%%%%%%%%%%%%%%%%%%%%%%%%%%%%%%%
\subsection{Diffusion Model}
\label{sec:method-DM}
%%%%%%%%%%%%%%%%%%%%%%%%%%%%%%%%%%%%%%%%%%%%

\begin{figure}[!htb]
    \centering
    \captionsetup{justification=centering}
    \includegraphics[width=0.99\linewidth]{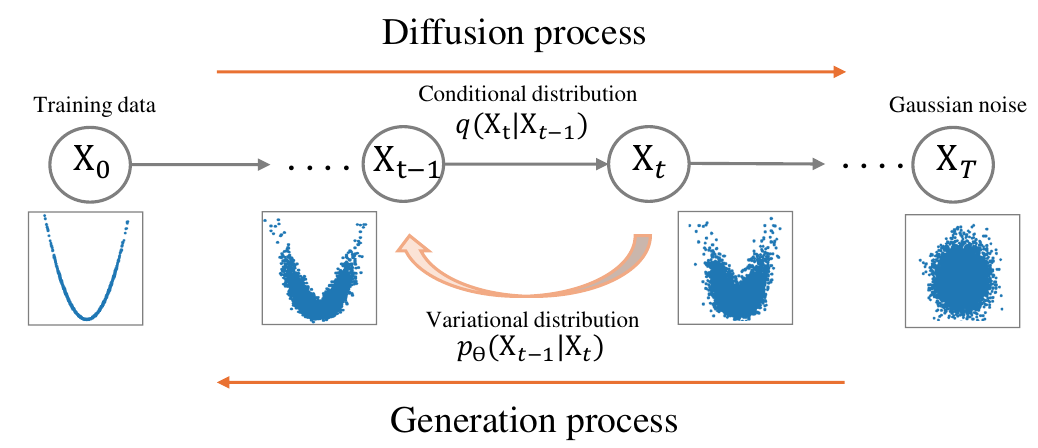}
    \caption{Illustration of the diffusion and generation processes of a DM.}
    \label{fig:DGMs-Diffusion-Models-Illustration}
\end{figure}

DMs draw inspiration from the principles of non-equilibrium thermodynamics \cite{sohl2015deep}. The training process of these models encompasses two distinct phases: the diffusion, or ``noising'' process, and the generation, or ``denoising'' process, as illustrated in Figure \ref{fig:DGMs-Diffusion-Models-Illustration}. During training, the model learns the underlying data distribution by successively adding Gaussian noise to the training data. Next, the model learns to reverse this noising process, aiming to accurately recover the original data \cite{ho2020denoising}. After training, DMs can be utilized to generate new samples simply by feeding random noise through the learned denoising process, as shown in Figure \ref{fig:DGMs-Diffusion-Models-Illustration}.

Given a data point $\mathbf{x}_0$ sampled from a real data distribution $\mathbf{x}_0 \sim q(\mathbf{x})$, in the forward diffusion process, small amounts of Gaussian noise are added to sample $\mathbf{x}_0$ in $T$ steps (also referred to as time-steps), producing a sequence of noisy samples $\{\mathbf{x}_t\}_{t=1}^T$. To avoid confusion with the ``time steps'' in transient modeling and simulation problems, in this work we will refer to $T$ as the number of ``time-steps''.
This process is done using a sequential transition kernel $q(\mathbf{x}_t \vert \mathbf{x}_{t-1})$, formulated as follows:
\begin{equation} \label{eqn:sequential-transition-kernel}
	q(\mathbf{x}_t \vert \mathbf{x}_{t-1}) = \mathcal{N}(\mathbf{x}_t; \sqrt{1 - \beta_t} \mathbf{x}_{t-1}, \beta_t \mathbf{I})
\end{equation}

where $\{\beta_t\}_{t=1}^T \in (0, 1)$ denotes the variance schedule, which controls the amount of noise added at each time-step and is also referred to as the step size. The original data sample $\mathbf{x}_0$ gradually loses its characteristics as the step $t$ becomes larger. When $T \to \infty$, $\mathbf{x}_T$ is equivalent to an isotropic Gaussian distribution. In real practice, we usually do not need a very large $T$ to arrive at this approximation.

The diffusion process can be collectively defined through a series of transition kernels:
\begin{equation} \label{eqn:forward-diffusion}
    q(\mathbf{x}_{1:T} \vert \mathbf{x}_0) = \prod^T_{t=1} q(\mathbf{x}_t \vert \mathbf{x}_{t-1}).
\end{equation}

Once the diffusion process is completed, we can recreate the true samples from a Gaussian noise input, $\mathbf{x}_T \sim \mathcal{N}(\mathbf{0}, \mathbf{I})$, as long as we can reverse the forward diffusion process and sample from $q(\mathbf{x}_{t-1} \vert \mathbf{x}_t)$. However, this cannot be easily done because we do not know $q(\mathbf{x}_{t-1} \vert \mathbf{x}_t)$. Therefore, we choose to learn a model $p_\theta (\mathbf{x}_{t-1} \vert \mathbf{x}_t)$ to approximate these conditional probabilities in order to run the reverse diffusion process. 
The generation process entails removing noise at each step in the reverse direction until the original training data is reconstructed. Starting with pure Gaussian noise $p(\mathbf{x_T}) = \mathcal{N}(\mathbf{x}_T; 0, \mathbf{I})$, the model learns the joint distribution $p_{\theta}(\mathbf{x_{0:T}})$:
\begin{equation}
	p_\theta(\mathbf{x}_{0:T}) 
	= p(\mathbf{x}_T) \prod^T_{t=1} p_\theta(\mathbf{x}_{t-1} \vert \mathbf{x}_t)
\end{equation}
where $\theta$ denotes the learnable parameters. By assuming the conditional distributions are approximately Gaussian, we have
\begin{equation}
	p_\theta (\mathbf{x}_{t-1} \vert \mathbf{x}_t) 
	= \mathcal{N} \left( \mathbf{x}_{t-1}; \bm{\mu}_\theta(\mathbf{x}_t, t), \bm{\Sigma}_\theta(\mathbf{x}_t, t) \right)
\end{equation}
where the mean vector $\bm{\mu}_\theta(\mathbf{x}_t, t)$ and the covariance matrix $\bm{\Sigma}_{\theta}(\mathbf{x}_t, t)$ are parameterized by deep neural networks. 

The training setup for a DM is very similar to a VAE, because $p_\theta (\mathbf{x}_{t-1} \vert \mathbf{x}_t)$ can be treated as a variational distribution for $q(\mathbf{x}_{t-1} \vert \mathbf{x}_t)$.
The simplified loss function, as derived by Ho et al. \cite{ho2020denoising} is used to train the DM:
\begin{equation}\label{eq:simplified-DM-loss}
    	\mathcal{L}_t= \mathbb{E}_{t \sim [1, T], \mathbf{x}_0, \bm{\epsilon}_t} \left[ \|\bm{\epsilon}_t - \bm{\epsilon}_\theta(\sqrt{\bar{\alpha}_t}\mathbf{x}_0 + \sqrt{1 - \bar{\alpha}_t}\bm{\epsilon}_t, t)\|^2 \right] 
\end{equation}
where $\bm{\epsilon}_t$ is the true noise added to the data, and $\bm{\epsilon}_\theta$ is the noise predicted by the model. The loss function is the mean squared error between the true and predicted noise. All symbols used in this section are defined in Table \ref{table:DM_symbols}.

\begin{table}[!htb] 
	\footnotesize
	\captionsetup{justification=centering}
	\caption{Definitions of mathematical symbols for DM.}
	\label{table:DM_symbols}
	\centering
	\begin{tabular}{cl}
		\toprule
	    Symbols & Meanings \\ 
		\midrule
        $q(\mathbf{x})$ & Real data distribution\\
        $\mathbf{x}_0$ & Data point sampled from real data distribution \\
        $\mathbf{x}_t$ & Noised latent variable at time-step $t$ \\
         $\bm{\mu}_\theta(\mathbf{x}_t, t)$ & Mean function at time-step $t$ \\
         $\bm{\Sigma}_{\theta}(\mathbf{x}_t, t)$& Covariance function at time-step $t$\\
         
        $\beta_t$ & Variance schedule  \\
        %$\alpha_t$ & $1 - \beta_t$ \\
        $\bm{\epsilon}_t$ & Gaussian noise at time-step $t$ \\
        %$\mathcal{L}_\text{VB}$ & Variational Bound (VB) loss function \\
        %$\mathcal{D}_{\text{KL}}$ & Kullback-Leibler divergence \\
        $\theta$ & Learnable parameters \\
        $\mathcal{L}_t$ &Divergence between inferred state at $\mathbf{x}_t$ and the true distribution at $t$ \\
        $\epsilon_{\theta}$ & Function approximator to predict $\epsilon$ from $\mathbf{x}_t$\\
		\bottomrule
	\end{tabular}
\end{table}

%%%%%%%%%%%%%%%%%%%%%%%%%%%%%%%%%%%%%%%%%%%%
\subsection{Conditional Diffusion Model}
\label{sec:method-CDM}
%%%%%%%%%%%%%%%%%%%%%%%%%%%%%%%%%%%%%%%%%%%%

The vanilla DM can only generate data randomly rather than under specific conditions. To address this issue, we developed a CDM that is capable of targeted synthetic data generation at user-provided TH conditions. A CDM performs similarly to a DM, starting with the successive addition of Gaussian noise. In the denoising process, the model learns to reverse the noising process to recover the original data. However, in a CDM, the denoising process is conditioned on additional data, guiding the model to generate data under the specified conditions. Once the model is trained, new data can be generated by passing a random vector along with a ``condition vector'' to the denoising direction of the model. 

By modifying the DM loss function from Equation (\ref{eq:simplified-DM-loss}), the loss function used to train a CDM can be expressed as follows \cite{dhariwal2021diffusion,park2025multi}: 

\begin{equation} \label{eq:simplified-CDM-loss}
    \mathcal{L}_t= \mathbb{E}_{t \sim [1, T], \mathbf{x}_0, \bm{\epsilon}_t} \left[ \|\bm{\epsilon}_t - \bm{\epsilon}_\theta(\sqrt{\bar{\alpha}_t}\mathbf{x}_0 + \sqrt{1 - \bar{\alpha}_t}\bm{\epsilon}_t, t,\mathbf{c})\|^2 \right] .
\end{equation}

The loss function is again the mean squared error between true and predicted noise. The $\bm{\epsilon}_\theta$ term is the noise predicted by the model. In this case, the model predicts noise based on the given input while incorporating the condition vector $\mathbf{c}$.

Using CDMs can be highly beneficial when a direct comparison between real and generated data is needed as they enable the direct calculation of errors and the estimation of uncertainties. During the generation process, the model takes a random vector along with user-provided TH conditions, which are from a reserved test dataset not seen in the training process. By changing the random vector while keeping the conditions fixed, different generated values can be obtained. This variability helps in estimating the model's uncertainty, thereby establishing the user's confidence in the generated values.

%%%%%%%%%%%%%%%%%%%%%%%%%%%%%%%%%%%%%%%%%%%%%%%%%%%%%%%%%%%%%%%%%%%%%%%%%%%%%%%%
\section{Problem Definition}
\label{sec:problem-definition}
%%%%%%%%%%%%%%%%%%%%%%%%%%%%%%%%%%%%%%%%%%%%%%%%%%%%%%%%%%%%%%%%%%%%%%%%%%%%%%%%

This section presents the problem definition. We will first describe the CHF phenomena and explain why it is crucial for nuclear energy, followed by an overview of the CHF training dataset in Section \ref{sec:problem-definition-CHF}. Sections \ref{sec:problem-definition-DM} and \ref{sec:problem-definition-CDM} include the training details of the DM and CDM models, respectively.

%%%%%%%%%%%%%%%%%%%%%%%%%%%%%%%%%%%%%%%%%%%%
\subsection{The CHF Dataset}
\label{sec:problem-definition-CHF}
%%%%%%%%%%%%%%%%%%%%%%%%%%%%%%%%%%%%%%%%%%%%

The goal of this study is to explore the capabilities of DM and CDM in generating realistic CHF data. CHF is one of the most important safety-related limiting quantities in nuclear energy. In the study of heat transfer in nuclear reactors, CHF consists of two different physical phenomena, departure from nucleate boiling which occurs mainly in pressurized water reactors with low quality flows, and dryout which occurs in high quality conditions typically found in boiling water reactors. In nuclear reactor operations, exceeding the CHF limit must be avoided because it can potentially lead to fuel rod failure, because it indicates that the heated fuel rod surface has reached a point where it can no longer efficiently transfer heat to the surrounding fluid. Therefore, data augmentation to expand the CHF experimental database is essential for the design and safety analysis of advanced water-cooled reactors.

The DM and CDM are trained on the largest known publicly available CHF experimental dataset, curated and published by the US NRC \cite{groeneveld2019critical}. The dataset consists of 24,579 CHF measurement data points in vertical uniformly-heated water-cooled tubes. It was compiled from 59 different experimental measurements performed during a span of 60 years, based on various CHF identification methods, such as visual identification, physical burnout, changes in the test section resistances, and the usage of thermocouples. The parameter space covered by this CHF database is substantial as it spans a wide range of commercial nuclear reactor operational conditions. It is worth noting that recently the same dataset has been used for an international benchmark on ``Artificial Intelligence and Machine Learning for Scientific Computing in Nuclear Engineering'' \cite{lecorre2024benchmark}, organized by the Organisation for Economic Co-operation and Development (OECD) Nuclear Energy Agency (NEA). The benchmark tasks include feature analysis and regression-based ML models development for CHF prediction. Our work on synthetic CHF dataset generation can significantly expand the size and diversity of the available CHF data, and thus improve the performances of such predictive ML models.

\begin{figure}[!htb]
    \centering
    \captionsetup{justification=centering}
    \includegraphics[width=0.95\linewidth]{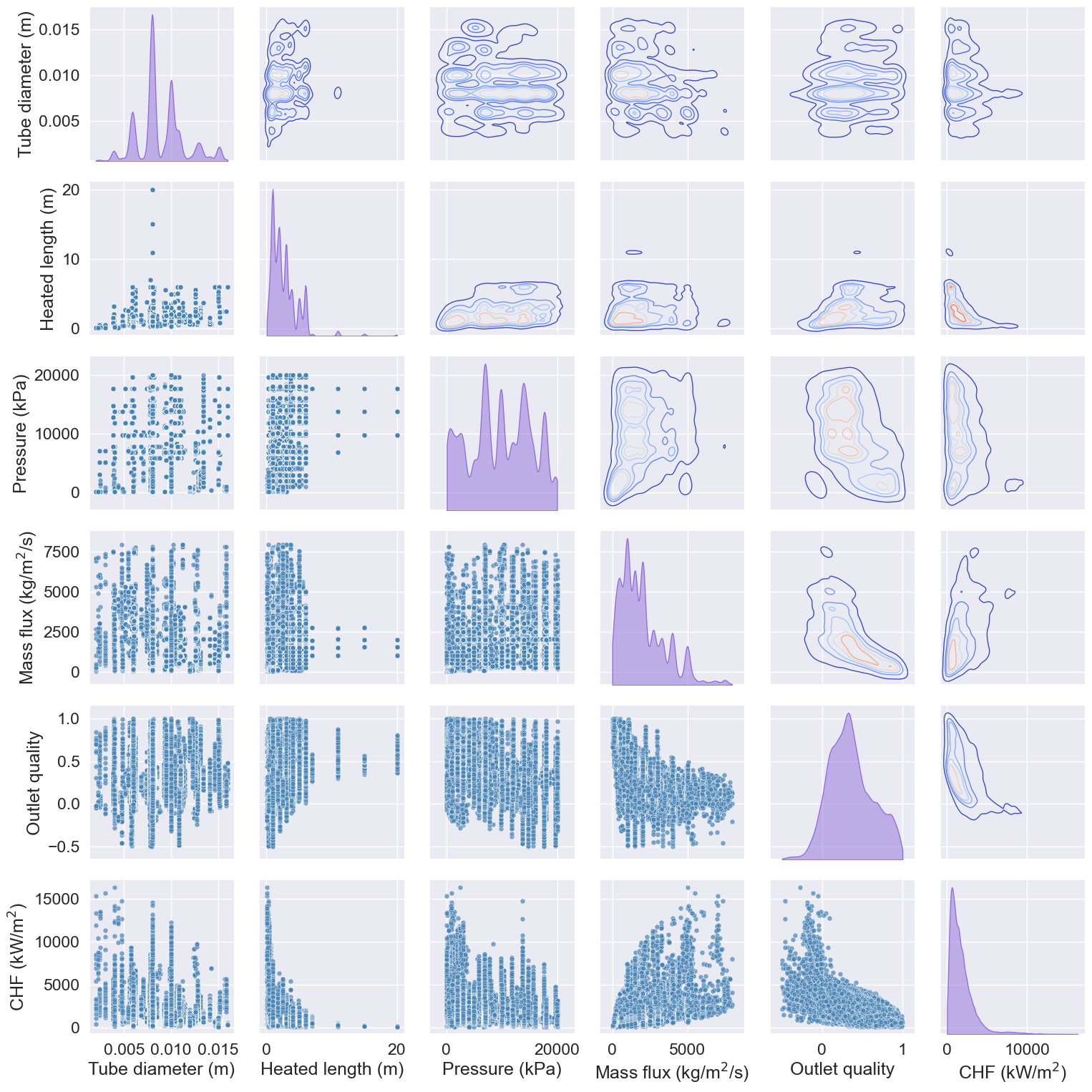}
    \caption{The distributions and correlations of the TH parameters and CHF values in the NRC CHF dataset.}
    \label{fig:CHF-data-plot-5-params}
\end{figure}

The CHF training dataset includes various TH parameters (also referred to as the features), including initial and boundary conditions such as pressure ($P$), mass flux ($G$), and inlet temperature ($T_{\text{in}}$), as well as geometrical parameters like test section diameter ($D$) and heated length ($L$). It also contains derived parameters from measurements and water properties, including outlet equilibrium quality ($x$) and inlet subcooling ($h_{\text{sub}}$). The distributions of these parameters, along with their pair-wise correlations, are shown in Figure \ref{fig:CHF-data-plot-5-params}. In this work, we have selected five TH conditions as the features $P$, $G$, $D$, $L$, and $x$, along with CHF, for the generative learning analysis. This is consistent with the recommendations from the OECD/NEA CHF benchmark summary \cite{lecorre2025oecd}, as predictive ML models with these input features have the best performance.

%%%%%%%%%%%%%%%%%%%%%%%%%%%%%%%%%%%%%%%%%%%%
\subsection{Training of DM}
\label{sec:problem-definition-DM}
%%%%%%%%%%%%%%%%%%%%%%%%%%%%%%%%%%%%%%%%%%%%

Training of a DGM is essentially an unsupervised ML process, because one treats the inputs (in this work, the five-dimensional TH conditions, $P$, $G$, $D$, $L$, and $x$) and outputs (the CHF values corresponding to the five-dimensional TH conditions) in a training sample as a vector, without learning the functional mapping from inputs to outputs. The DGM learns the joint distributions of all these parameters together. Once trained, the synthetic samples being generated also have the same dimension, with the intention that the CHF-TH-parameter relationship can be maintained for such an unsupervised learning process. For DMs, random noise is passed during the generation process, and the model produces samples consisting of five TH parameters and the corresponding CHF value. The major limitation in a vanilla DM is that we cannot directly evaluate accuracy of the generated CHF values by holding out a blind test dataset, because the synthetic samples are randomly generated. Therefore, the vanilla DM will be evaluated based on performance by assessing the following:
\begin{enumerate}
    \item The ability to capture the distributions of each TH parameter and the CHF output. This is done by comparing the marginal distributions of the true and generated samples.
    \vspace{-0.5em}
    \item The ability to capture the correlations between parameters in the dataset, evaluated by comparing the Pearson correlation coefficient (PCC) and Spearman rank correlation coefficient (SRCC) of the measurement and generated datasets.
    \vspace{-0.5em}
    \item The generation of physically meaningful values for all parameters.
    \vspace{-0.5em}
    \item The overall distributional similarity between the generated data and the training data, assessed by comparing the full joint empirical cumulative distribution functions (ECDFs) of the real and synthetic datasets. 
\end{enumerate}

These metrics can evaluate if the model has learned the underlying distribution of the training data, and if the synthetic samples can be considered as drawn from the original distribution.

The DM was trained using the simplified loss function utilized by Ho et al. in the fundamental paper for denoising diffusion probabilistic models \cite{ho2020denoising} as shown in Equation (\ref{eq:simplified-DM-loss}). To stabilize training, the exponential moving average (EMA) technique was used. Instead of directly updating the model's parameters, a copy of the previous parameters is maintained and updated as a weighted mean between the current and previous values. This can help to improve the training stability by reducing the impact of individual parameter updates. EMA is calculated as a weighted combination of the current parameter values and their previous values:

\begin{equation}\label{eqn:EMA}
    \text{EMA} = (1-\mu) \times \text{parameter}_{\text{current}} + \mu \times \text{parameter}_{\text{previous}}
\end{equation}
where $\mu$ is the smoothing constant, with a value between 0 and 1. A lower value gives more weight to the current parameter, while a higher value gives more weight to the previous parameter. During the training of the DM, we used a value of 0.9, as suggested in \cite{ho2020denoising}. 

The model utilized $T=100$ time-steps for the diffusion process. Gaussian noise was added step-by-step using a sigmoid variance schedule ($\beta$), with noise values smoothly transitioning between $1 \times 10^{-5}$ and $1 \times 10^{-2}$. Smaller noise values were added at the beginning of the diffusion process, while larger noise values were added in the later time-steps, following the shape of the sigmoid curve. The noise continues to grow larger towards the end of the process as the sigmoid function approaches its upper bound. The DM was trained for 1,200 epochs with a batch size of 64 and a learning rate of $1 \times 10^{-3}$.

%%%%%%%%%%%%%%%%%%%%%%%%%%%%%%%%%%%%%%%%%%%%
\subsection{Training of CDM}
\label{sec:problem-definition-CDM}
%%%%%%%%%%%%%%%%%%%%%%%%%%%%%%%%%%%%%%%%%%%%

\begin{figure}[!htb]
    \centering
    \captionsetup{justification=centering}
    \includegraphics[width=0.99\linewidth]{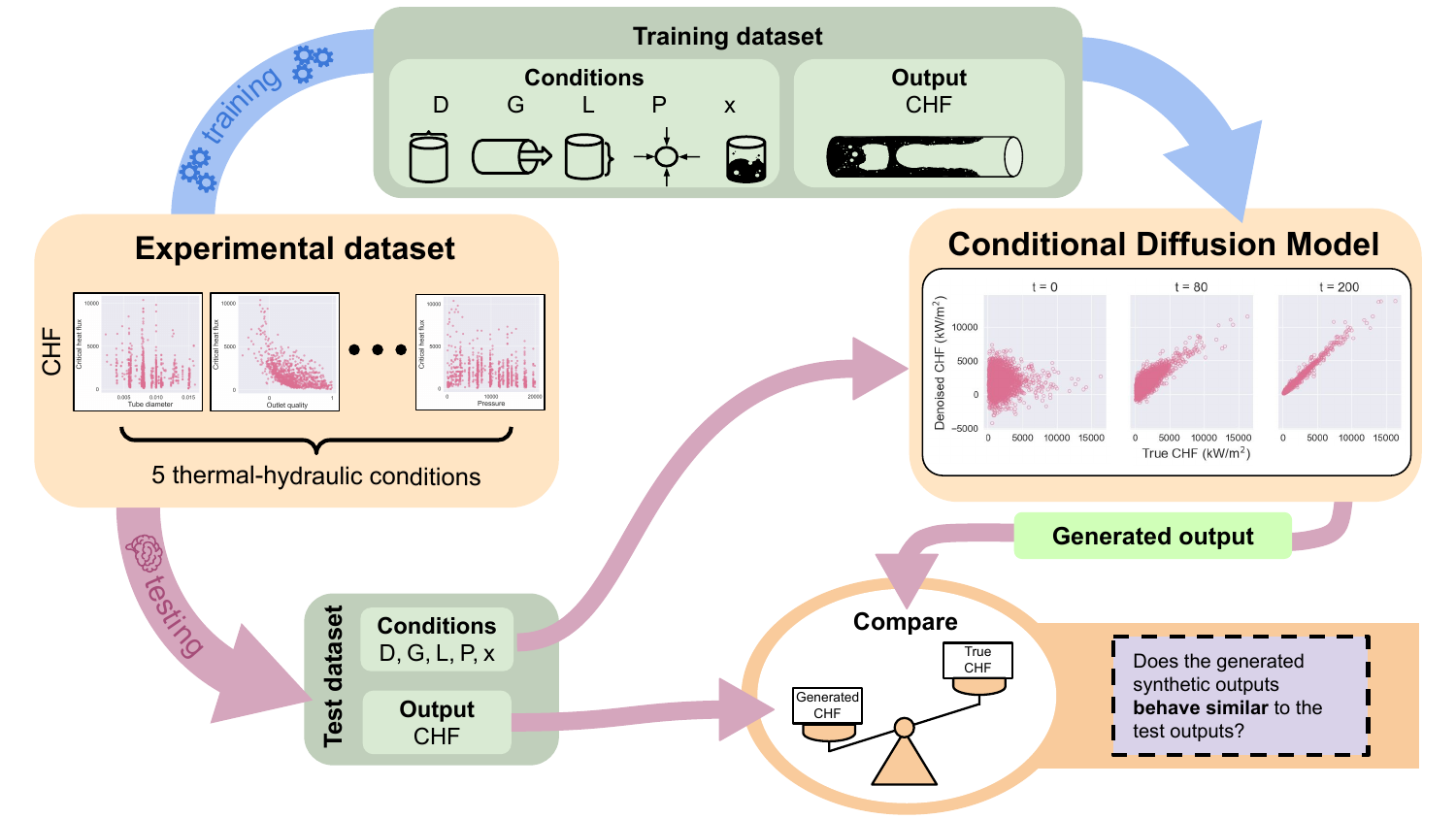}
    \caption{Flowchart outlining the process of CDM training.}
    \label{fig:CDM_flowchart_diagram}
\end{figure}

A CDM was trained to generate CHF values under user-specified TH conditions. These conditions consist of the five TH parameters: $P$, $G$, $D$, $L$, and $x$. The dataset is split into 80\% for training, 10\% for hyperparameter tuning, and 10\% for blind testing. By inputting TH condition values from the held-out test dataset, the CDM generates CHF values corresponding to these specified TH conditions. The model is then evaluated by a direct comparison with the true CHF values in the test dataset not seen by the model during the training or hyperparameter tuning steps. Figure \ref{fig:CDM_flowchart_diagram} illustrates the workflow for CDM training.

UQ analysis is performed by leveraging the inherent variability in the CDM generation process. Different samples of random noise fed to the isotropic Gaussian distribution, combined with the same TH conditions, will produce varying outputs through the generation process. To quantify uncertainties in the generated data, we repeat the generation process 500 times, providing a different random vector each time while keeping the conditions fixed. We then compute the mean of the 500 generated samples ($\mu_\text{samples}$) for each condition vector in the test dataset, along with the standard deviation ($\sigma_\text{samples}$), which provides an estimate of the uncertainties in the generated data.

The CDM was trained using the loss function presented in Equation (\ref{eq:simplified-CDM-loss}). To stabilize the training process, EMA was also applied, as discussed in Section \ref{sec:problem-definition-DM}. For this model, the EMA value was set to 0.9. The model architecture consisted of six hidden layers, and a learning rate of $10^{-4}$ was used. The model was trained for 7,500 epochs with a batch size of 300. Gaussian noise was added in each time-step using a sigmoid variance schedule ($\beta$) ranging from $10^{-5}$ to $10^{-2}$ over $T=200$ time-steps. Hyperparameter tuning was performed for the variance schedule, noise range, number of time-steps, and model-specific hyperparameters (e.g., number of hidden layers, batch size).

%%%%%%%%%%%%%%%%%%%%%%%%%%%%%%%%%%%%%%%%%%%%%%%%%%%%%%%%%%%%%%%%%%%%%%%%%%%%%%%%
\section{Results using DM}
\label{sec:results-DM}
%%%%%%%%%%%%%%%%%%%%%%%%%%%%%%%%%%%%%%%%%%%%%%%%%%%%%%%%%%%%%%%%%%%%%%%%%%%%%%%%

A DM was trained to augment the CHF dataset by taking random Gaussian noise and generating synthetic samples, each consisting of values for the five TH parameters and the corresponding CHF. In this work, 10,000 synthetic samples were generated for the assessment of the DM. Note that one can generate an arbitrary number of new samples at an increased cost, however, one cannot control what samples will be generated. That is, the vanilla DM will generate new samples randomly, rather than at specific TH conditions provided by the user. As a result, we cannot evaluate its generative accuracy by a direct comparison with real CHF values. Instead, we compared the collective statistical behavior of the generated samples with the real experimental data.

\begin{figure}[!htb]
    \centering
    \captionsetup{justification=centering}
    \includegraphics[width=\linewidth]{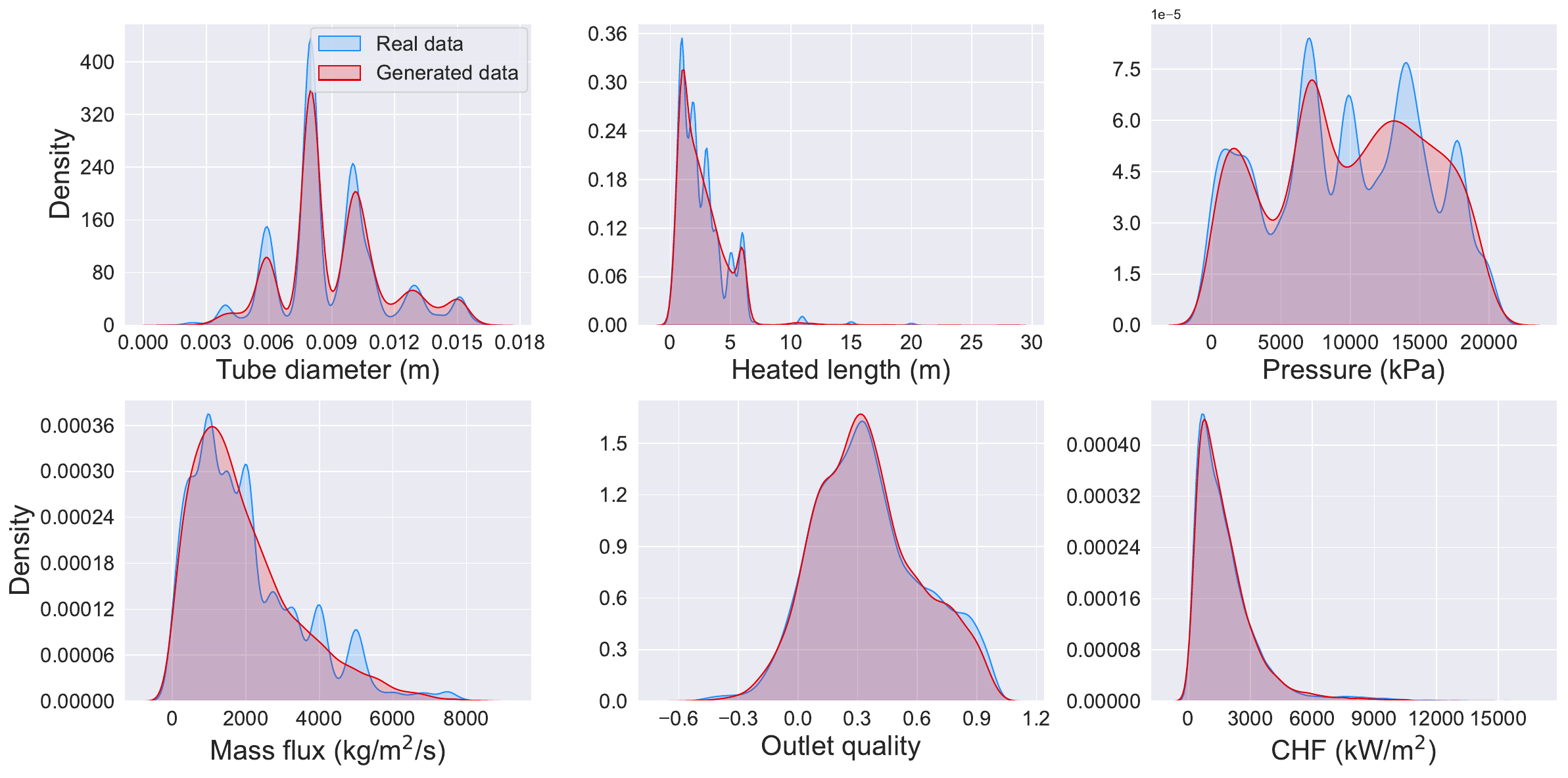}
    \caption{Comparison of the TH parameters and CHF distributions between the real and DM-generated data. }
    \label{fig:DM-CHF-real-vs-generated-distributions-all-parameters}
\end{figure}

Figure \ref{fig:DM-CHF-real-vs-generated-distributions-all-parameters} shows a comparison of the distributions from kernel density estimation of the five TH parameters and CHF, between the real CHF dataset and the DM-generated dataset. The distributions of some TH parameters were found to be smoother in the synthetic data. For instance, in the case of tube diameter ($D$), the measurement data is derived from experiments with discrete values for $D$, leaving gaps for values where no experiments were conducted. The generated data contains $D$ values that do not exist in the measurement data, while closely following the true distribution. This observation also applies to the heated length, pressure and mass flux. The outlet quality and the CHF values have similar smooth distributions between the two datasets, this is mainly because the values in the training dataset have a relatively continuous coverage of the parameter domains.

\begin{figure}[!htb]
   \centering
   \captionsetup{justification=centering}
   \includegraphics[width=\linewidth]{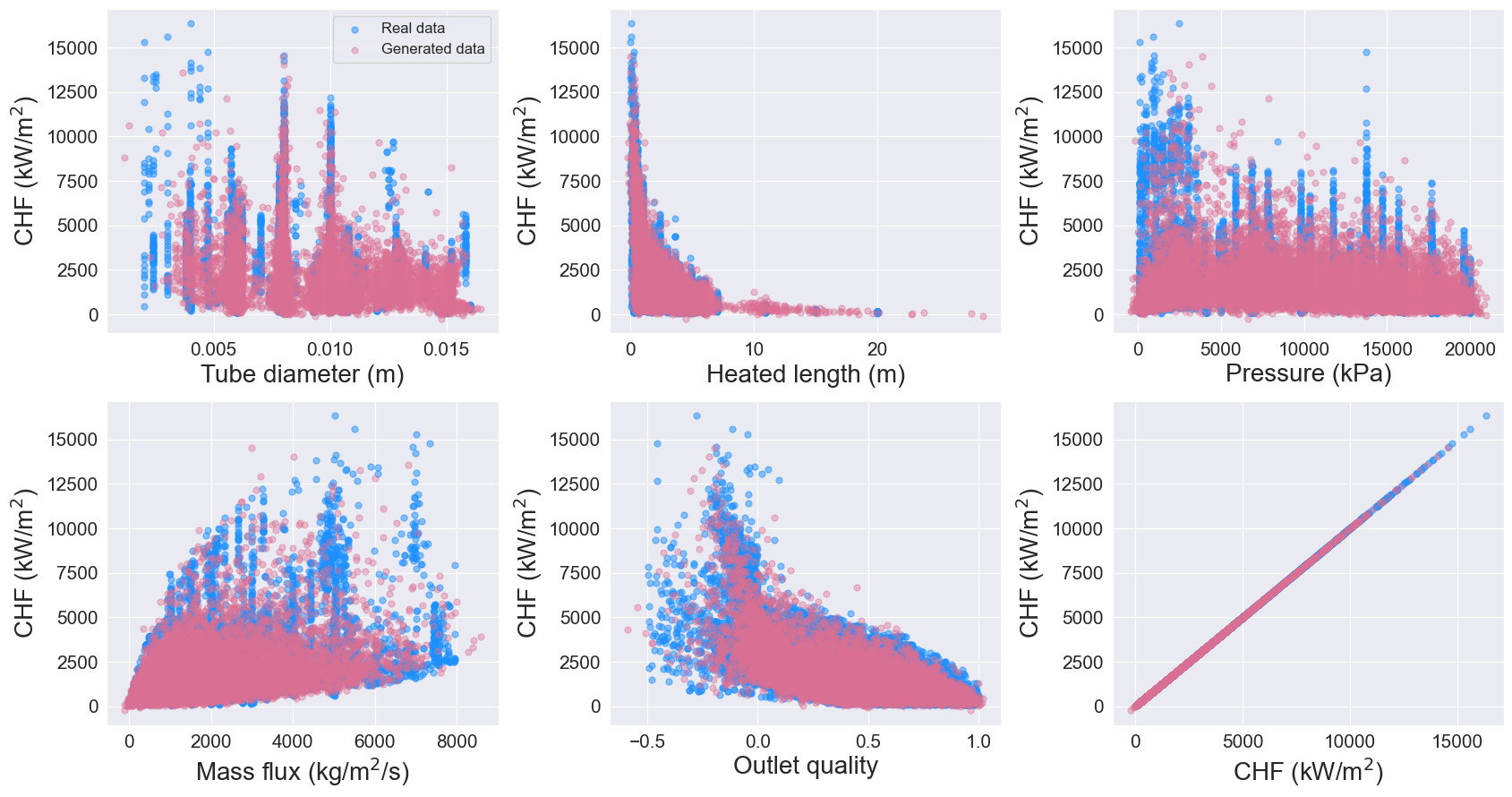}
   \caption{Comparison of the CHF-TH-parameter pairwise correlations between the real and DM-generated data. }
    \label{fig:DM-CHF-real-vs-generated-pairwise-correlations-all-parameters}
\end{figure}

To assess whether the model successfully learned the correlations between CHF and the TH parameters in the real dataset, and can subsequently generate values while preserving these correlations, Figure \ref{fig:DM-CHF-real-vs-generated-pairwise-correlations-all-parameters} shows a visual comparison of the CHF-TH-parameter pairwise correlations between the real data and the data generated by the DM. The generated data notably preserves the correlations between CHF and the TH parameters when compared to the real data, while also producing new samples not present in the training dataset. 

\begin{figure}[!htb]
    \centering
    \captionsetup{justification=centering}
    \includegraphics[width=0.8\linewidth]{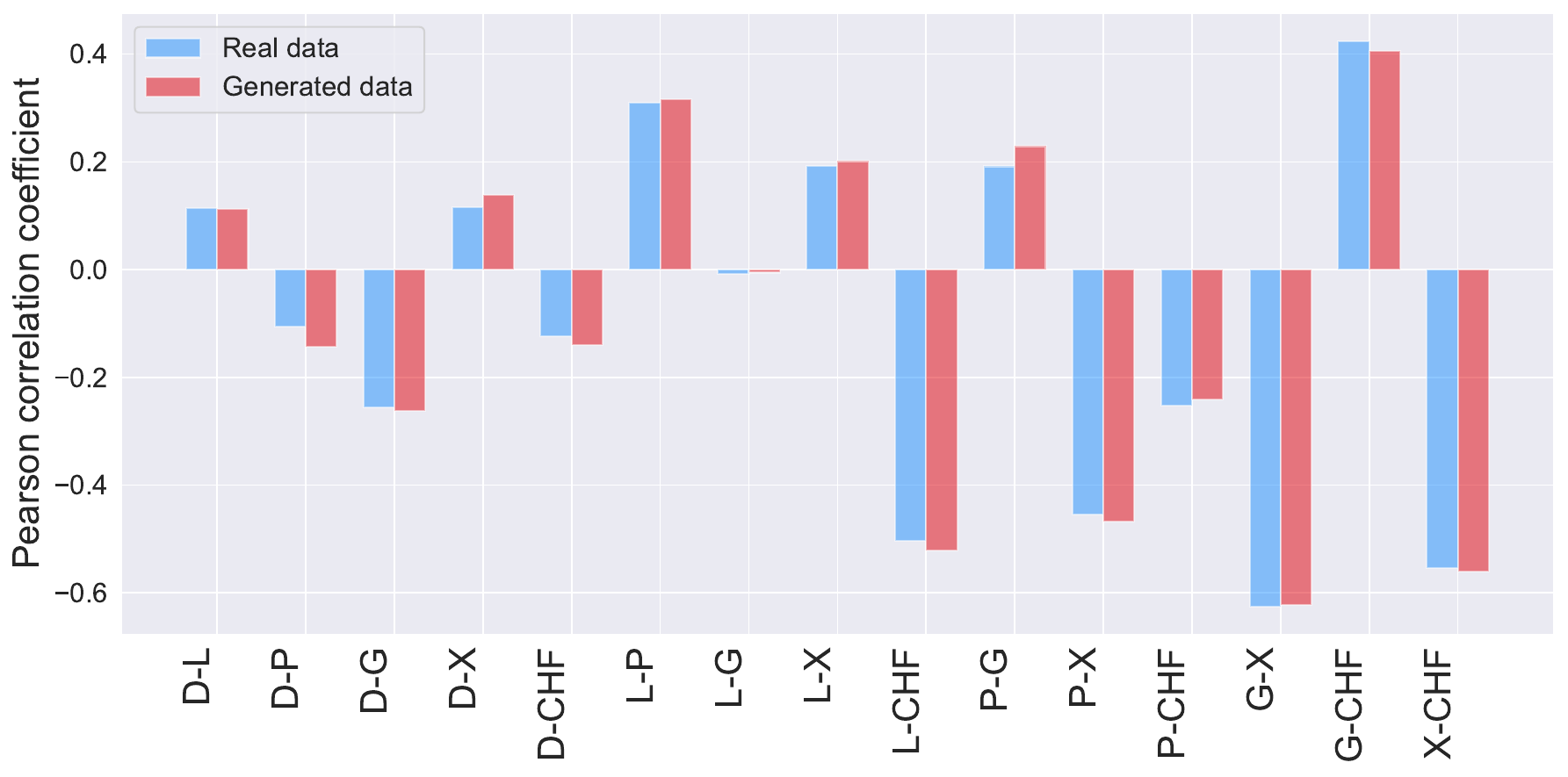}
    \caption{Comparison of PCCs of the CHF-TH-parameters between the real and DM-generated data.}
    \label{fig:DM-CHF-Comparison-of-Pearson-Correlation-Coefficients}
\end{figure}

\begin{figure}[!htb]
    \centering
    \captionsetup{justification=centering}
    \includegraphics[width=0.8\linewidth]{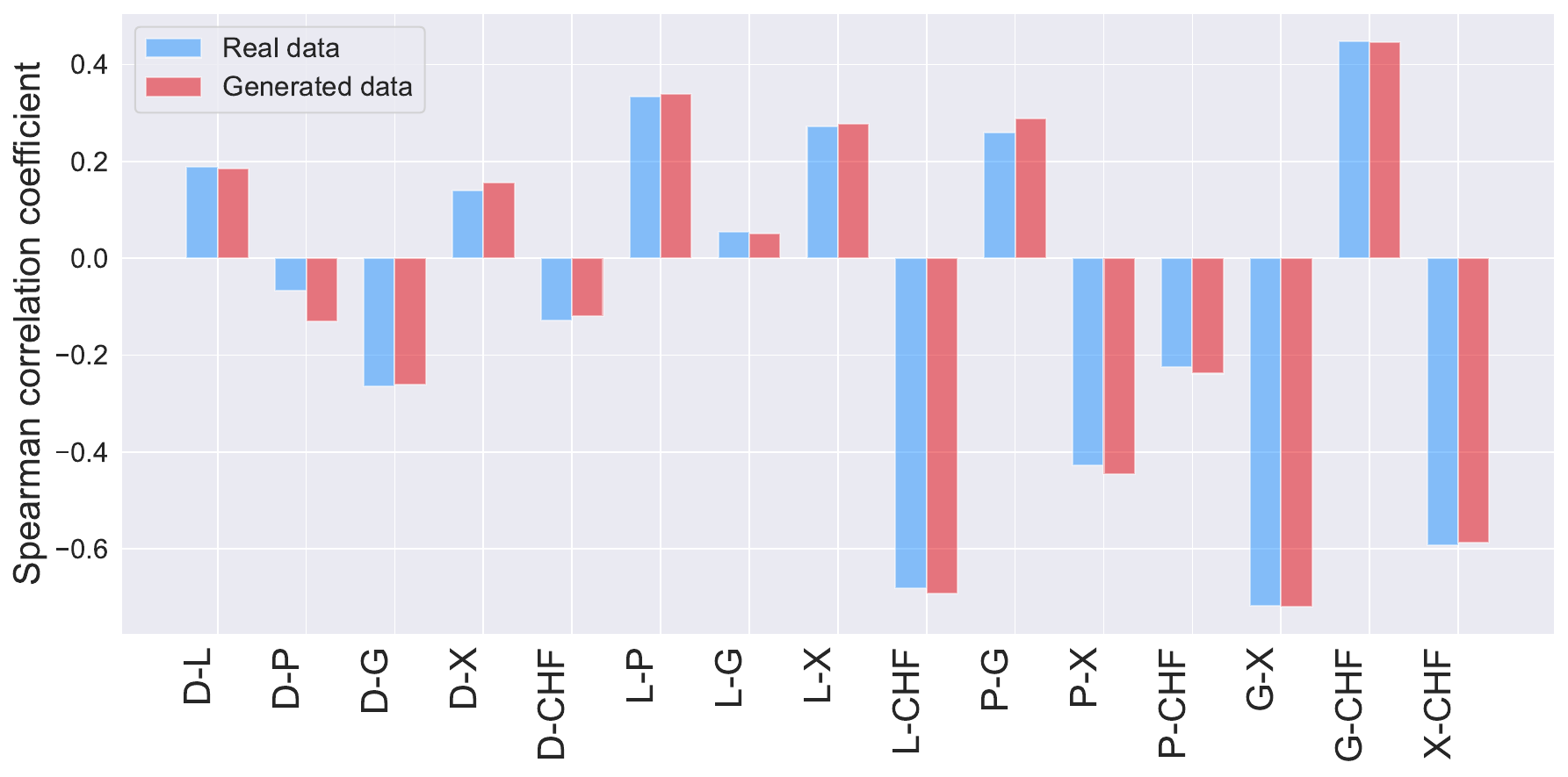}
    \caption{Comparison of SRCCs of the CHF-TH-parameters between the real and DM-generated data. }
    \label{fig:DM-CHF-Comparison-of-Spearman-Correlation-Coefficients}
\end{figure}

In addition to the visual comparison shown in Figure \ref{fig:DM-CHF-real-vs-generated-pairwise-correlations-all-parameters}, we also calculated the pairwise PCCs/SRCCs between all the five TH parameters and CHF in the training data, and compared each correlation metric with the corresponding values from the generated data. The PCC captures linear relationships, while the SRCC assesses monotonic relationships, which can be either linear or non-linear. Both metrics were used to evaluate whether the model successfully learned the correlations between the six variables (5 TH parameters and CHF). The comparisons between the PCC values and the SRCC values are shown in Figures \ref{fig:DM-CHF-Comparison-of-Pearson-Correlation-Coefficients} and \ref{fig:DM-CHF-Comparison-of-Spearman-Correlation-Coefficients}, respectively. The values indicate that the DM was able to capture the correlations between the parameters, as the PCCs and SRCCs calculated from the training data agree well with those from the generated data. Small deviations are observed in some pairs of variables; however, these deviations are insignificant.

\begin{figure}[!htb]
    \centering
    \includegraphics[width=0.99\linewidth]{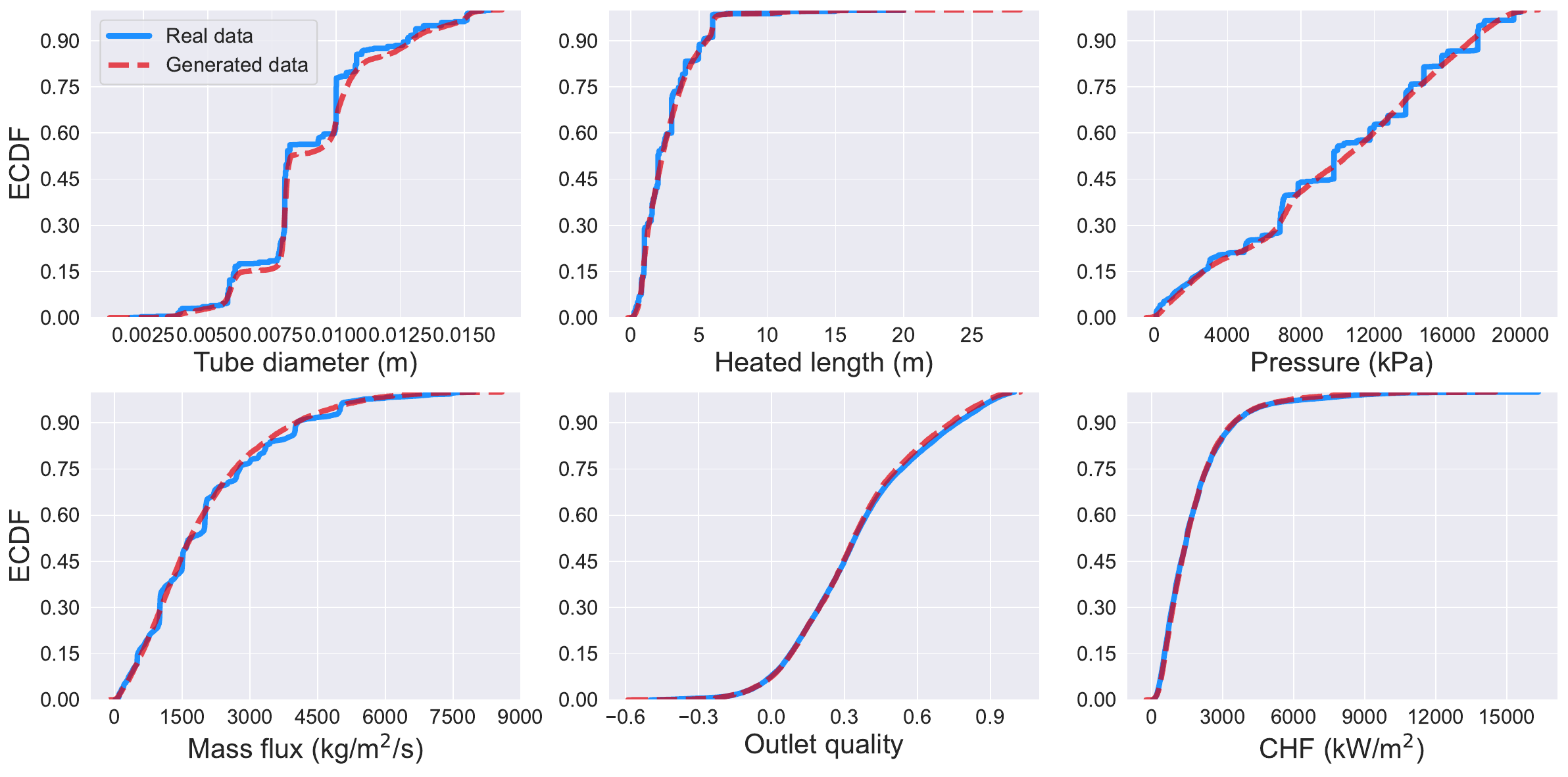}
    \caption{Comparison of the six marginal ECDFs between the real and DM-generated data.}
    \label{fig:DM-CHF-real-vs-generated-ECDFs}
\end{figure}

The DM was also evaluated by testing the overall distributional similarity between the generated data and the real data by comparing the ECDFs of the six variables (five TH parameters + CHF). A comparison of the ECDFs between the real data and DM-generated data in Figure \ref{fig:DM-CHF-real-vs-generated-ECDFs} shows a strong agreement between the ECDFs.

In addition to a visual inspection of the six marginal ECDFs in Figure \ref{fig:DM-CHF-real-vs-generated-ECDFs}, we utilized the genai-evaluation library for a quantitative comparison of the joint ECDF of all the six variables. This library was designed specifically for evaluating the quality of tabular synthetic data. First, we calculated the multivariate ECDFs of both datasets. Afterwards, we compared the similarity between the real and generated data by computing the Kolmogorov-Smirnov (KS) distance between the two ECDFs. This metric produces a value between 0 and 1, where values close to 0 indicate that the two distributions are similar, while a value of 1 corresponds to distributions that are completely dissimilar. 
The KS distance between the two joint ECDFs was found to be 0.1265, indicating that the joint ECDF of the generated data is similar to that of the training data. This indicates that the DM learned the distribution of the training data and generated realistic synthetic samples.

%%%%%%%%%%%%%%%%%%%%%%%%%%%%%%%%%%%%%%%%%%%%%%%%%%%%%%%%%%%%%%%%%%%%%%%%%%%%%%%%
\section{Results using CDM}
\label{sec:results-CDM}
%%%%%%%%%%%%%%%%%%%%%%%%%%%%%%%%%%%%%%%%%%%%%%%%%%%%%%%%%%%%%%%%%%%%%%%%%%%%%%%%

%%%%%%%%%%%%%%%%%%%%%%%%%%%%%%%%%%%%%%%%
\subsection{Results of CHF Generation}
\label{sec:results-CDM-CHF-generation}
%%%%%%%%%%%%%%%%%%%%%%%%%%%%%%%%%%%%%%%%

In this subsection, we present the results of the CDM in generating CHF values. The major difference with the vanilla DM is that the CDM can generated targeted CHF values at user-provided TH conditions. In this work, 10\% of the real CHF dataset has been reserved for testing. Evaluating the trained CDM at the same TH conditions with the test dataset will generate CHF values corresponding to these TH conditions. This enables us to compare the real and synthetic CHF values directly to evaluate the accuracy of the generated CHF samples, which was not possible for the vanilla DM in Section \ref{sec:results-DM}. Therefore, in this subsection, we will present some extra results using quantitative metrics.

\begin{figure}[!htb]
    \centering
    \includegraphics[width=\linewidth]{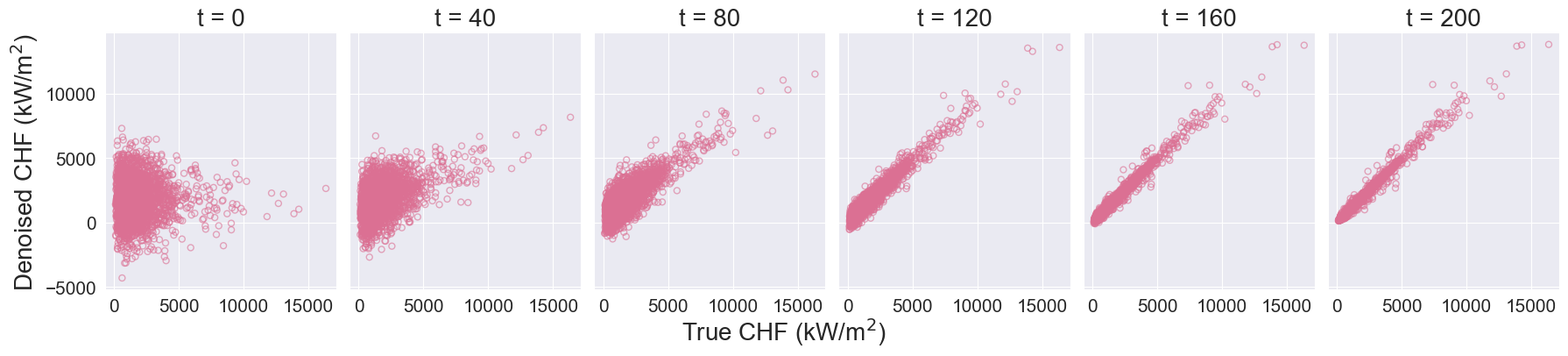}
    \caption{The denoising process of CDMs, starting from random noise at $t = 0$ and gradually removing noise to generate the final CHF values at $t = 200$.}
    \label{fig:CDM-denoising-process}
\end{figure}

Figure \ref{fig:CDM-denoising-process} presents a visualization of the denoising process, showing the generation of CHF data under the TH conditions from the testing dataset. It illustrates how the model gradually removes noise over 200 time-steps, starting from pure Gaussian noise at $t=0$ and ending with the generated CHF values at $t=200$. The figure displays the denoised samples every 40 time-steps against the true CHF values on the x-axis. Note that the noise and generated CHF values are shown after being rescaled, as the data was standardized using a standard scaler before being fed into the model. 

\begin{figure}[!htb]
    \centering
        \begin{subfigure}{0.495\textwidth}
        \includegraphics[width=0.99\textwidth]{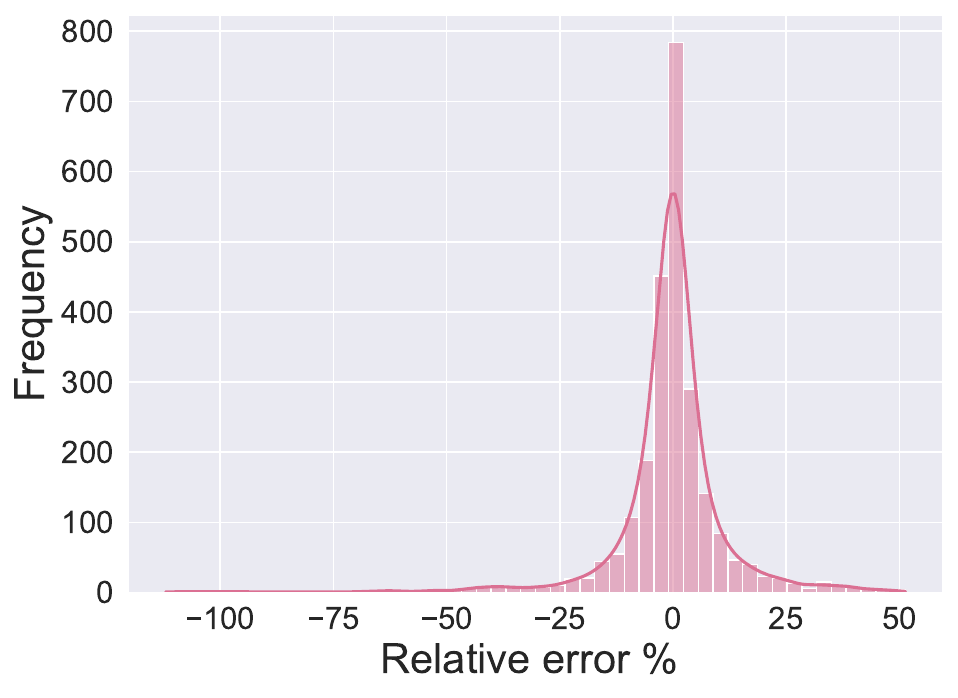}
        \caption{Relative error distribution}
        \label{fig:CDM-error-dist}
    \end{subfigure}
    \begin{subfigure}{0.495\textwidth}
        \includegraphics[width=0.99\textwidth]{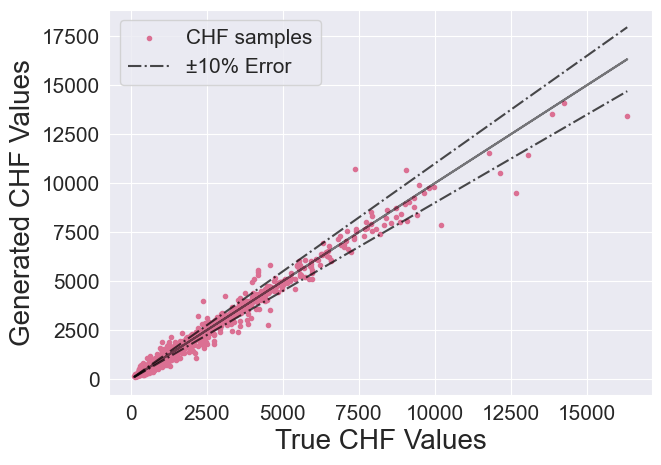}
        \caption{True vs. generated CHF values}
        \label{fig:CDM-true-vs-generated}
    \end{subfigure}
    \caption{Performance assessment of CDM based on a direct comparison of the true vs. CDM-generated CHF values.}
    \label{fig:CDM-model-performance}
\end{figure}

The model's performance was evaluated by directly comparing the generated CHF values with the true CHF values from the held-out testing dataset. The relative error between the generated and true CHF values was calculated, and its distribution is presented in Figure \ref{fig:CDM-error-dist}. The majority of errors in the generated samples fall within $\pm 25\%$, with most testing points exhibiting small error values centered around the mean relative error of around 6.8\%. Additionally, less than 5.8\% of the testing data show errors exceeding $\pm 25\%$. 

Figure \ref{fig:CDM-true-vs-generated} presents a parity plot of the generated and true CHF values, including the $\pm10\%$ error bounds. The generated CHF values show strong agreement with the true CHF values. To better interpret the results, several statistical metrics on the absolute relative errors, are presented in Table \ref{table:metrics-RE}. These metrics include the mean, maximum, and standard deviation of the absolute relative errors, as well as the fraction of testing points with an absolute relative error exceeding 10\%. The results show that the CDM accurately generated CHF values, with a mean absolute relative error of 6.8\%. While the maximum error was significantly higher than the mean, it occurred at very small CHF values, where small deviations from the true value can lead to a high relative error. Notably, only 5.8\% of the data had errors greater than 25\%, while 19.3\% had errors greater than 10\%. Furthermore, the $R^2$ value of approximately 0.98 indicates a strong correlation between the true and generated values. 

\begin{table}[!htb] 
	\normalsize
	\captionsetup{justification=centering}
	\caption{Statistical metrics for the absolute relative errors between the real CHF values and those generated by CDM for the testing dataset.}
	\label{table:metrics-RE}
	\centering
	\begin{tabular}{crr}
        \toprule
        Statistical metrics & Base case (Section \ref{sec:results-CDM-CHF-generation}) & With UQ (Section \ref{sec:results-CDM-UQ})\\
        \midrule
        $\mu_{\text{error}}$ & 6.86 \%  & 5.89 \% \\
        %\hline
        $\text{Max}_{\text{error}}$ & 111.84 \%  &113.55   \% \\
        %\hline
        $\text{Std}_{\text{error}}$ & 10.64 \%  &8.66  \% \\
        %\hline
        $F_{\text{error}}>10\%$ & 19.32 \%  &  18.18 \% \\
        \bottomrule
	\end{tabular}
\end{table}

\begin{figure}[!htb]
    \centering
    \includegraphics[width=\linewidth]{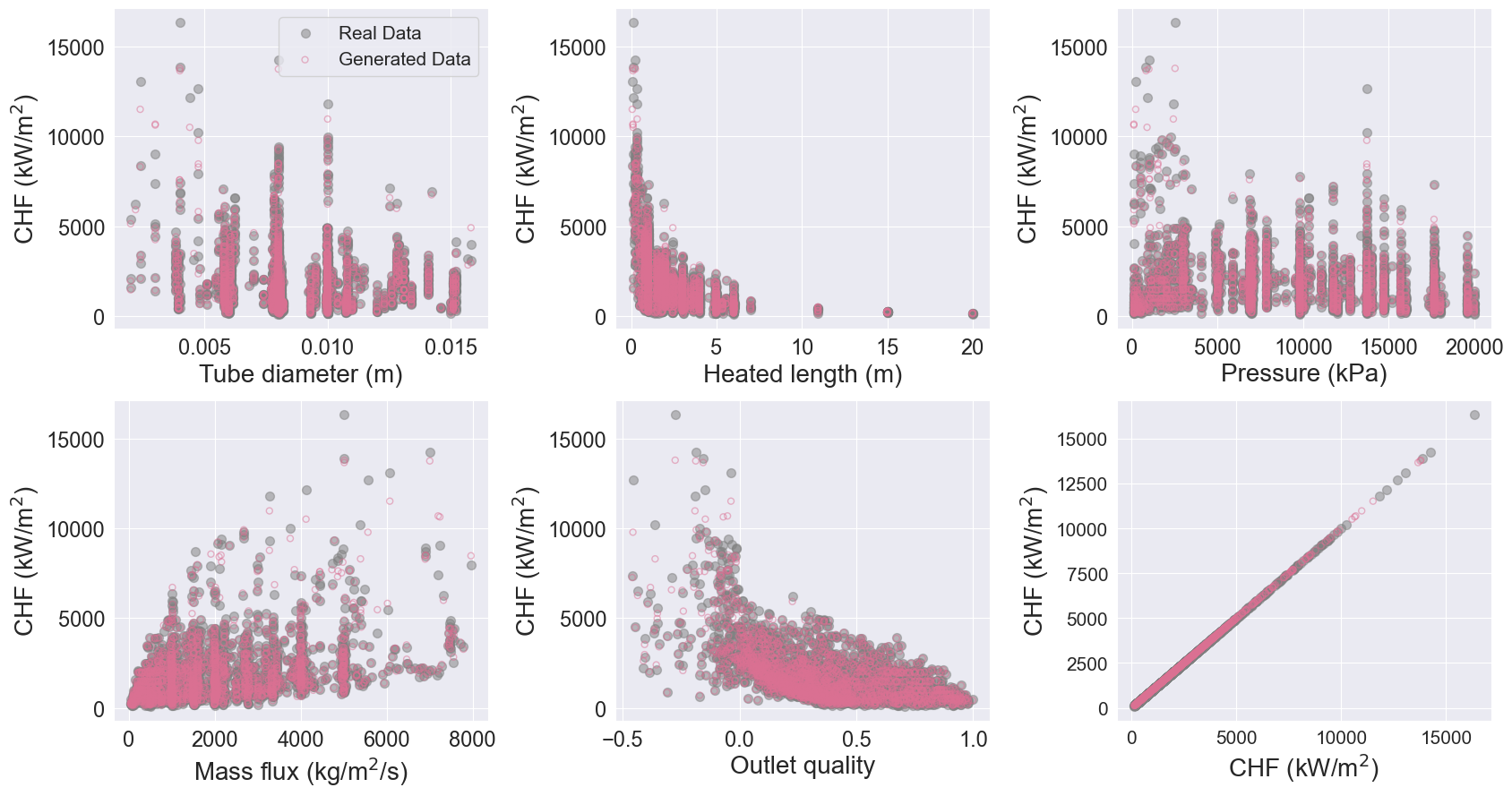}
    \caption{Comparison of the CHF-TH-parameter pairwise correlations between the real data and the CDM-generated data.}
    \label{fig:CDM-CHF-real-vs-generated-pairwise-correlations-all-parameters}
\end{figure}

Figure \ref{fig:CDM-CHF-real-vs-generated-pairwise-correlations-all-parameters} presents the pairwise correlations between CHF and the TH parameters for both the CDM-generated and real data. The figure demonstrates a strong alignment between the real and generated CHF values. To avoid redundancy, equivalent results to those presented in Section \ref{sec:results-DM} will not be shown here. Overall, these results prove that the CDM performs well in generating CHF values based on the testing dataset.

%%%%%%%%%%%%%%%%%%%%%%%%%%%%%%%%%%%%%%%%
\subsection{UQ Analysis}
\label{sec:results-CDM-UQ}
%%%%%%%%%%%%%%%%%%%%%%%%%%%%%%%%%%%%%%%%

Having established the CDM's generative performance, we next evaluate its uncertainty behavior. UQ is essential for assessing the reliability and establishing confidence in the generated data. UQ of CDM was performed as discussed in Section \ref{sec:problem-definition-CDM}. Using the trained CDM, 500 samples were generated for each data point in the testing dataset. In other words, for each TH condition defined by $P$, $G$, $D$, $L$, and $x$, the generation process of the CDM was evaluated 500 times to get an ensemble of 500 CHF values, which serve as Monte Carlo samples for the CHF at the chosen TH condition. The randomness comes from the generation of Gaussian noise at $t=0$ in the generation step.
The mean ($\mu_\text{samples}$) and standard deviation ($\sigma_\text{samples}$) of these samples were then computed, and the relative standard deviation was calculated using the Equation (\ref{eqn:rel_std}) to allow for better comparison across different scales of outputs.
\begin{equation} \label{eqn:rel_std}
    \text{Relative Std (\%)} = \frac{\sigma_{\text{samples}}}{\mu_{\text{samples}}} \times 100\%
\end{equation}

\begin{figure}[!htb] 
    \centering
    \includegraphics[width=0.6\linewidth]{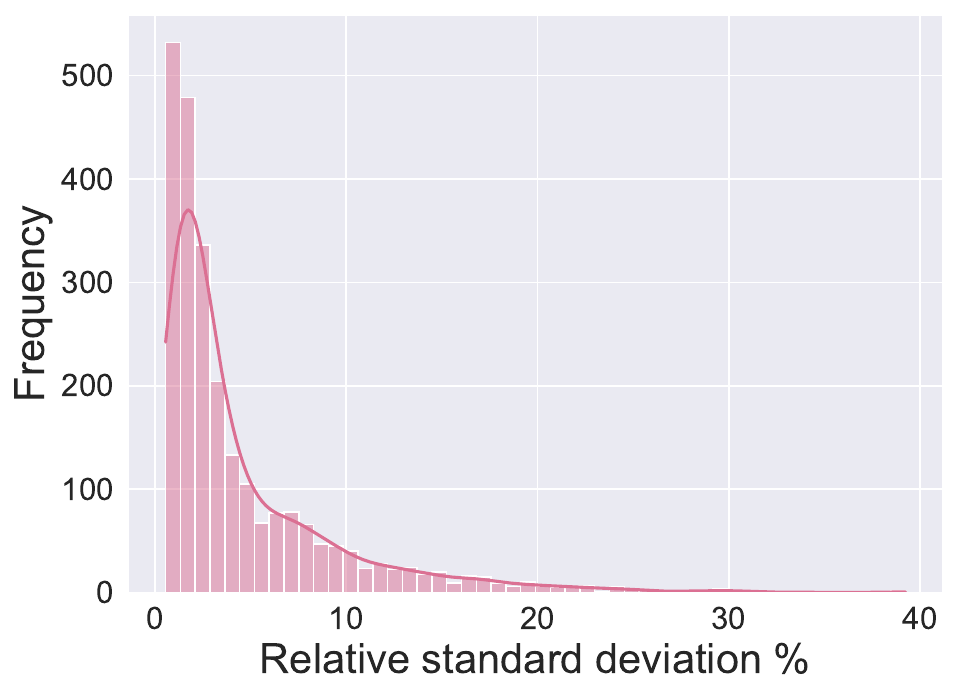}
    \caption{Distribution of the relative standard deviations. }
    \label{fig:CHF-CDM-UQ-Rstd-DLPGX}
\end{figure}

Figure \ref{fig:CHF-CDM-UQ-Rstd-DLPGX} presents the distribution of relative standard deviation values across the testing dataset. The majority of relative standard deviation values are concentrated around the mean value of 4.40\%, with most values remaining below 10\%. This indicates that the CDM produces relatively stable predictions with low variability. Only a few data points exhibit significantly higher relative standard deviation values, with a maximum of 44.47\%, suggesting the presence of instances where the model has higher variability in the generated values.
Again, we would like to point out that many CHF values in the testing dataset are very small, as shown in Figure \ref{fig:CDM-CHF-real-vs-generated-pairwise-correlations-all-parameters}. For these cases, a small standard deviation value can result in a large relative standard deviation.

Table \ref{table:metrics-RE} shows the same error metrics used in Section \ref{sec:results-CDM-CHF-generation}, with the errors calculated by comparing $\mu_{\text{samples}}$ with the true CHF value. The results from Table \ref{table:metrics-RE} also show improvement in the $\mu_{\text{error}}$ values dropping from 6.85\% to 5.89\%, when compared to the base case (without UQ), indicating slightly increased accuracy in CDM-generated values. This was also the case for the other metrics except for the $\text{Max}_{\text{error}}$, which slightly increased.

%%%%%%%%%%%%%%%%%%%%%%%%%%%%%%%%%%%%%%%%
\subsection{Validation of Physical Consistency}
\label{sec:results-CDM-physical-consistency}
%%%%%%%%%%%%%%%%%%%%%%%%%%%%%%%%%%%%%%%%

In order to use the synthetic data for downstream ML-based predictive tasks, a validation of the physical consistency is required. The assessments performed in Sections \ref{sec:results-DM} and \ref{sec:results-CDM-CHF-generation} were mostly based on comparison of the statistical behavior of the real and generated datasets. Although there is a sizable literature on physics-informed ML (sometimes referred to as ``physics-aware'', ``physics-constrained'',  ``physics-guided'', etc.), the methodology for embedding physics information in the training process is problem-dependent, so is the assessment of physical consistency. In this section, we propose an approach to validate the physical consistency of the generated data using the available information from the public CHF dataset, as well as the physical dependence among the TH features.

For the CHF dataset, we propose to use theoretical and empirical mathematical models underlying the CHF phenomena that describe the physical dependence among the TH parameters. The outlet equilibrium quality ($x$) can be obtained via the outlet enthalpy using an energy balance equation  \cite{hall1999ultra}, as well as by using steam tables to derive the saturated liquid and vapor enthalpies. In this work the XSteam library was used when the steam tables were needed. Equation (\ref{eq:LeCorre_Equations}) shows the process to obtain the calculated outlet equilibrium quality ($x$) using other TH parameters and the CHF.

\begin{equation}
	\begin{aligned} \label{eq:LeCorre_Equations}
		h_f &= h_\text{L}(P) \\
		h_{fg} &= h_\text{V}(P) - h_f \\
		h_{\text{in}} &= h_f - h_{\text{sub}} \\
		\Delta h &= \frac{4\,q_{\text{CHF}}\,L}{G\,D} \\
		h_{\text{out}} &= h_{\text{in}} + \Delta h \\
		x_{\text{calc}} &= \frac{h_{\text{out}} - h_f}{h_{fg}}
	\end{aligned}
\end{equation}

In brief, to check the physical consistency in the synthetic samples generated by CDM, we compare the outlet equilibrium quality ($x$) from the following sources:
\begin{itemize}
    \item ``Measured'': $x$ is available from the public CHF dataset. It has been used as a TH condition for the DM and CDM models in Sections \ref{sec:results-DM}, \ref{sec:results-CDM-CHF-generation}, and \ref{sec:results-CDM-UQ}. Note that, $x$ is not directly measured but derived based on CHF and other TH variable measurements and water properties, as mentioned in Section \ref{sec:problem-definition-CHF}. It is referred to as ``measured'' here because it is available from the public CHF dataset, and to avoid confusion with "calculated" $x$ below.
    \item ``Calculated'': $x$ can be calculated using other TH features together with the CHF values in the measurement data using Equation (\ref{eq:LeCorre_Equations}). Note that there will be some discrepancies between the ``Measured'' and ``Calculated'' values for $x$ due to two major reasons. Firstly, transcription errors have been common in CHF experimental dataset caused by either human mistakes or technological limitations, considering the long history of some of the CHF datasets. Secondly, many historical CHF measurements used older thermodynamic properties of water to derive the outlet equilibrium quality ($x$) and inlet subcooling ($h_{\text{sub}}$), for example, based on previous versions of the International Association for the Properties of Water and Steam (IAPWS) libraries. As a result, the newly calculated values for $x$ will be different from those in the public CHF dataset.
    \item ``Generated'': $x$ can be calculated using generated data from a CDM that is conditioned on $D$, $G$, $P$, $L$, and $h_\text{sub}$, using Equation (\ref{eq:LeCorre_Equations}). Note that this is a new CDM that is different from the CDM used in Sections \ref{sec:results-CDM-CHF-generation}, and \ref{sec:results-CDM-UQ}, which was conditioned on $D$, $G$, $P$, $L$, and $x$. 
\end{itemize}

\begin{figure}[!htb]
    \centering
    \includegraphics[width=0.99\linewidth]{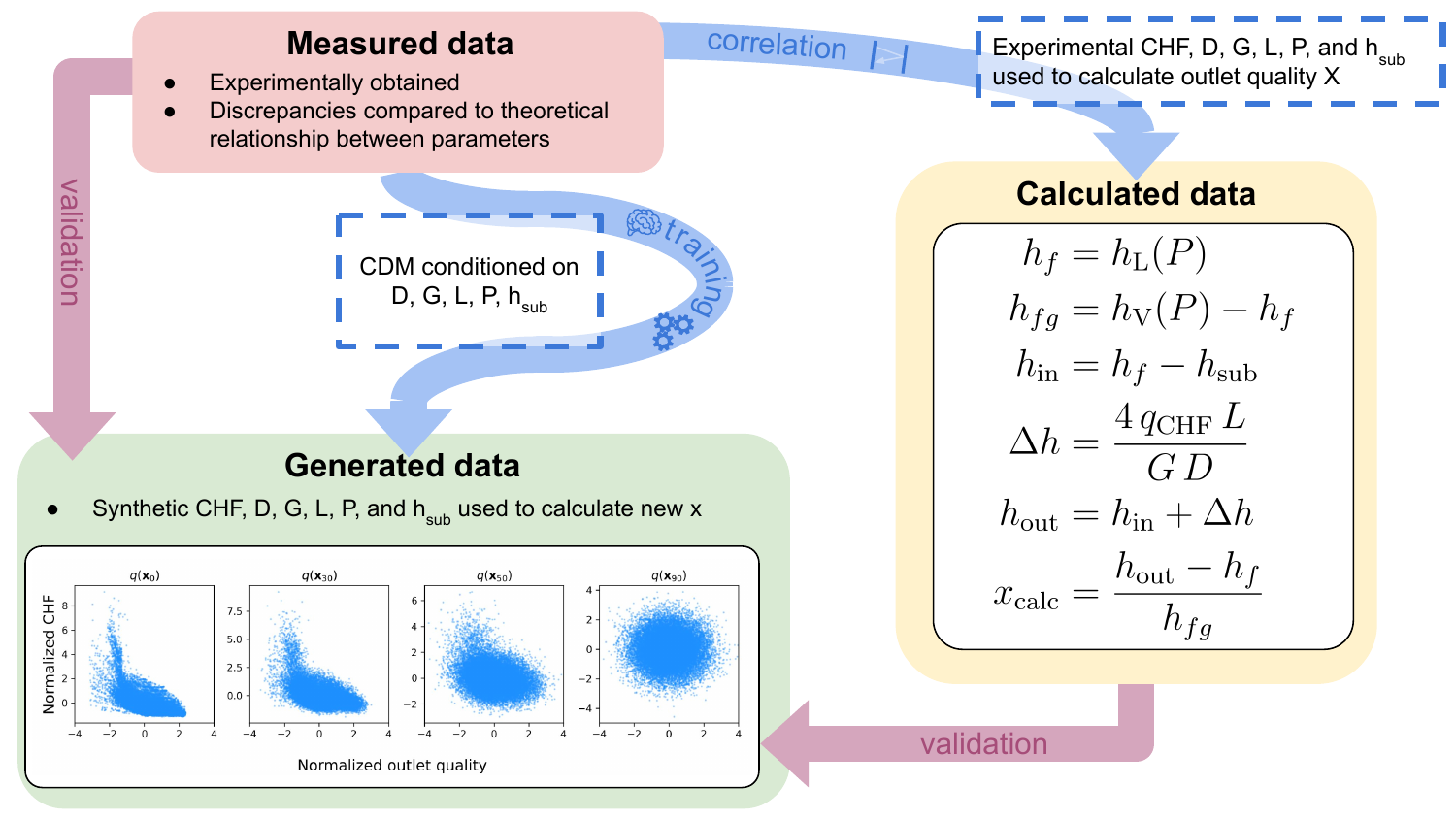}
    \caption{Workflow for physical consistency assessment of the outlet equilibrium quality ($x$).}
    \label{fig:CHF-CDM-Validation_flowchart}
\end{figure}

To validate the physical consistency, we checked the absolute errors in $x$ by comparing ``measured - generated'' and ``calculated - generated'', as shown in Figure \ref{fig:CHF-CDM-Validation_flowchart}. The goal is to examine if the consistency in $x$ can be maintained with extra steps of calculation with Equation (\ref{eq:LeCorre_Equations}) and generation by the CDM. Note that the CDM has been changed to be conditioned on a new TH parameter $h_\text{sub}$ because we would like to calculate the outlet equilibrium quality ($x$) for comparison based on generated $h_\text{sub}$. The reason that we used $x$ instead of $h_\text{sub}$ in earlier analysis is to be consistent with the recommendations from the OECD/NEA CHF benchmark summary \cite{lecorre2025oecd}.

\begin{figure}[!htb] 
    \centering   
    \includegraphics[width=0.6\linewidth]{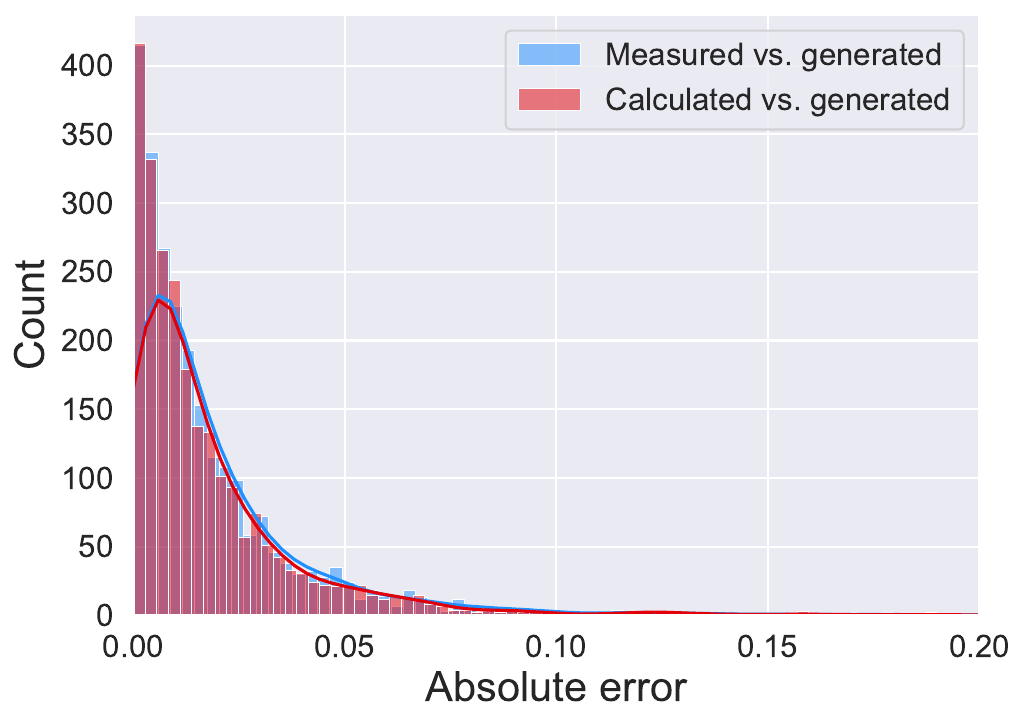}
    \caption{Comparison of absolute error distributions of the generated outlet equilibrium quality with measured and calculated values.}
    \label{fig:CHF-CDM-Validation_AE_histogram}
\end{figure}

For a physically consistent DGM, the generated data should perform the same when compared to both the measured and calculated data. The absolute error was selected as the metric on which to compare the discrepancies of the generated data, as a small portion of the $x$ values are very close to zero, which biases performance representation to any outliers. The results are summarized in Table \ref{tab:validation_absolute_error}. Performance of the generated qualities across the two comparisons are very close. The distributions of the absolute errors are visualized in Figure \ref{fig:CHF-CDM-Validation_AE_histogram}, which shows that the error distributions are very similar, and most absolute error values are small in magnitude.

\begin{table}[!htb] 
    \normalsize
    \captionsetup{justification=centering}
    \caption{Comparison of statistical metrics for the absolute errors.}
    \label{tab:validation_absolute_error}
    \centering
    \begin{tabular}{crr}
        \toprule
        Statistical metrics & Measured - generated & Calculated - generated \\
        \midrule
        Mean  & 0.0197 & 0.0190 \\
        %\hline
        STD   & 0.0292 & 0.0288 \\
        %\hline
        Min   & 7.0e-9  & 6.0e-6  \\
        %\hline
        25\%  & 0.0045 & 0.0043 \\
        %\hline
        50\%  & 0.0114 & 0.0107 \\
        %\hline
        75\%  & 0.0242 & 0.0229 \\
        %\hline
        Max   & 0.5823 & 0.5834 \\
        \bottomrule
    \end{tabular}
\end{table}

%%%%%%%%%%%%%%%%%%%%%%%%%%%%%%%%%%%%%%%%%%%%%%%%%%%%%%%%%%%%%%%%%%%%%%%%%%%%%%%%
\section{Summaries and Conclusions}
\label{sec:conclusions}
%%%%%%%%%%%%%%%%%%%%%%%%%%%%%%%%%%%%%%%%%%%%%%%%%%%%%%%%%%%%%%%%%%%%%%%%%%%%%%%%

Most applications of DMs have focused on text, image, or video data, and their applications for data augmentation of scientific datasets, particularly in the energy sector, have been relatively limited. There has been no existing work on DMs and CDMs for overcoming data scarcity in the nuclear energy domain. 
The main objective of this work is to investigate the effectiveness of DMs in generating physics-consistent data for CHF, which is one of the most important safety-related limiting quantities in nuclear energy. By leveraging a public dataset on CHF that covers a wide range of commercial nuclear reactor operational conditions, we developed a DM that can generate an arbitrary amount of synthetic samples for augmenting the CHF dataset. 
Statistical analysis of the DM results showed that the DM effectively learned the distribution of the training data and generated realistic samples. Additionally, the model was able to capture the correlations between the parameters in the dataset, producing a negligible number of unphysical samples.

Another goal of this work is to develop DGMs capable of targeted data generation.
Since a vanilla DM can only generate samples randomly, we also developed a CDM that can generate targeted CHF data at user-specified TH conditions. 
The performance of the CDM was evaluated through direct comparison with the CHF values in the held-out testing dataset. The results showed that the CDM successfully generated CHF data with small relative error values.

Finally, we developed a method to validate the physical consistency of the generated data using the available information from the public CHF dataset, as well as the physical dependence among the TH features. We used theoretical and empirical mathematical models underlying the CHF phenomena that describe the physical dependence among the TH parameters to prove that the CDM-generated values for the outlet equilibrium quality can maintain a good physical-consistency with CHF and the other TH parameters.

Even though the DM has been proven to be effective in synthetic data generation, it must be noted that the DM-generated samples can only cover the gaps in the existing experimental dataset to a relatively small extent, as shown in Figure \ref{fig:DM-CHF-real-vs-generated-pairwise-correlations-all-parameters}. The synthetic samples still tend to fall in the same ranges as the training dataset. The CDM is preferred as it is capable of targeted data generation. Future work is still needed for data augmentation in extrapolated domains beyond the training domain. Our future plan also includes combining deep generative modeling and transfer learning to overcome data scarcity in nuclear energy.

%%%%%%%%%%%%%%%%%%%%%%%%%%%%%%%%%%%%%%%%%%%%%%%%%%%%%%%%%%%%%%%%%%%%%%%%%%%%%%%%
\section*{Acknowledgement}
%%%%%%%%%%%%%%%%%%%%%%%%%%%%%%%%%%%%%%%%%%%%%%%%%%%%%%%%%%%%%%%%%%%%%%%%%%%%%%%%

\noindent This work was funded by the U.S. Department of Energy (DOE) Office of Nuclear Energy Distinguished Early Career Program (DECP) under award number DE-NE0009467. Any opinions, findings, and conclusions or recommendations expressed in this paper are those of the authors and do not necessarily reflect the views of the U.S. DOE.

%%%%%%%%%%%%%%%%%%%%%%%%%%%%%%%%%%%%%%%%%%%%%%%%%%%%%%%%%%%%%%%%%%%%%%%%%%%%%%%%
%%%%%%%%%%%%%%%%%%%%%%%%%%%%%%%%%%%%%%%%%%%%%%%%%%%%%%%%%%%%%%%%%%%%%%%%%%%%%%%%
%%\section*{References}
\bibliography{./bibliography.bib}

@techreport{groeneveld2019critical,
	title={Critical Heat Flux Data Used to Generate the 2006 Groeneveld Lookup Tables},
	author={Groeneveld, D. C.},
	year={2019},
	institution={NUREG/KM-0011, Office of Nuclear Regulatory Research, U.S. Nuclear Regulatory Commission (NRC)}
}

@article{groeneveld20072006,
	title={The 2006 CHF look-up table},
	author={Groeneveld, DC and Shan, JQ and Vasi{\'c}, AZ and Leung, LKH and Durmayaz, A and Yang, J and Cheng, SC and Tanase, A},
	journal={Nuclear engineering and design},
	volume={237},
	number={15-17},
	pages={1909--1922},
	year={2007},
	publisher={Elsevier}
}

@techreport{lecorre2024benchmark,
	title={Benchmark on Artificial Intelligence and Machine Learning for Scientific Computing in Nuclear Engineering. Phase 1: Critical Heat Flux Exercise Specifications},
	author={Le Corre, Jean-Marie and Delipei, Gregory and Wu, Xu and Zhao, Xingang},
	year={2024},
	institution={NEA Working Papers, NEA/WKP(2023)1, OECD Publishing, Paris}
}

@inproceedings{lecorre2025oecd,
	title={OECD/NEA Benchmark on Artificial Intelligence and Machine Learning for Critical Heat Flux Predictions – Summary of Phase 1 Results},
	author={Le-Corre, Jean-Marie and Delipei, Gregory and Zhao, Xingang and Wu, Xu and Buss, Oliver},
	booktitle={Proceedings of the 21\textsuperscript{st} International Topical Meeting on Nuclear Reactor Thermal Hydraulics (NURETH-21)},
	year={2025},
	organization={Busan, Republic of Korea, August 31 - September 5, 2025}
}

@article{hall1999ultra,
	title={Ultra-high critical heat flux (CHF) for subcooled water flow boiling—II: high-CHF database and design equations},
	author={Hall, David D and Mudawar, Issam},
	journal={International Journal of Heat and Mass Transfer},
	volume={42},
	number={8},
	pages={1429--1456},
	year={1999},
	publisher={Elsevier}
}

@article{alsafadi2023deep,
	title={Deep Generative Modeling-based Data Augmentation with Demonstration using the BFBT Benchmark Void Fraction Datasets},
	author={Alsafadi, Farah and Wu, Xu},
	journal={Nuclear Engineering and Design},
	volume={415},
	pages={112712},
	year={2023},
	publisher={Elsevier}
}

@article{alsafadi2025predicting,
	title={Predicting critical heat flux with uncertainty quantification and domain generalization using conditional variational autoencoders and deep neural networks},
	author={Alsafadi, Farah and Furlong, Aidan and Wu, Xu},
	journal={Annals of Nuclear Energy},
	volume={220},
	pages={111502},
	year={2025},
	publisher={Elsevier}
}

@article{alsafadi2025investigation,
	title={An Investigation on Machine Learning Predictive Accuracy Improvement and Uncertainty Reduction using VAE-based Data Augmentation},
	author={Alsafadi, Farah and Yaseen, Mahmoud and Wu, Xu},
	journal={Nuclear Engineering and Design},
	volume={445},
	pages={114433},
	year={2025},
	publisher={Elsevier}
}

@article{liu2025unsupervised,
	title={Unsupervised anomaly detection for Nuclear Power Plants based on Denoising Diffusion Probabilistic Models},
	author={Liu, Shiqiao and Zhu, Zifei and Zhao, Xinwen and Wang, Yangguang and Sun, Xiang and Yu, Lei},
	journal={Progress in Nuclear Energy},
	volume={178},
	pages={105521},
	year={2025},
	publisher={Elsevier}
}

@article{salakhutdinov2015learning,
	title={Learning deep generative models},
	author={Salakhutdinov, Ruslan},
	journal={Annual Review of Statistics and Its Application},
	volume={2},
	number={1},
	pages={361--385},
	year={2015},
	publisher={Annual Reviews}
}

@article{ruthotto2021introduction,
	title={An introduction to deep generative modeling},
	author={Ruthotto, Lars and Haber, Eldad},
	journal={GAMM-Mitteilungen},
	volume={44},
	number={2},
	pages={e202100008},
	year={2021},
	publisher={Wiley Online Library}
}

@article{mirza2014conditional,
  title={Conditional generative adversarial nets},
  author={Mirza, Mehdi and Osindero, Simon},
  journal={arXiv preprint arXiv:1411.1784},
  year={2014}
}

@article{goodfellow2020generative,
  title={Generative adversarial networks},
  author={Goodfellow, Ian and Pouget-Abadie, Jean and Mirza, Mehdi and Xu, Bing and Warde-Farley, David and Ozair, Sherjil and Courville, Aaron and Bengio, Yoshua},
  journal={Communications of the ACM},
  volume={63},
  number={11},
  pages={139--144},
  year={2020},
  publisher={ACM New York, NY, USA}
}

@article{baasch2021conditional,
	title={A Conditional Generative adversarial Network for energy use in multiple buildings using scarce data},
	author={Baasch, Gaby and Rousseau, Guillaume and Evins, Ralph},
	journal={Energy and AI},
	volume={5},
	pages={100087},
	year={2021},
	publisher={Elsevier}
}

@article{rizzato2022stress,
	title={Stress testing electrical grids: Generative adversarial networks for load scenario generation},
	author={Rizzato, Matteo and Morizet, Nicolas and Mar{\'e}chal, William and Geissler, Christophe},
	journal={Energy and AI},
	volume={9},
	pages={100177},
	year={2022},
	publisher={Elsevier}
}

@article{carreon2023generative,
	title={A generative adversarial network (GAN) approach to creating synthetic flame images from experimental data},
	author={Carreon, Anthony and Barwey, Shivam and Raman, Venkat},
	journal={Energy and AI},
	volume={13},
	pages={100238},
	year={2023},
	publisher={Elsevier}
}

@article{zhang2023solargan,
	title={SolarGAN: Synthetic annual solar irradiance time series on urban building facades via Deep Generative Networks},
	author={Zhang, Yufei and Schlueter, Arno and Waibel, Christoph},
	journal={Energy and AI},
	volume={12},
	pages={100223},
	year={2023},
	publisher={Elsevier}
}

@article{walter2024probabilistic,
	title={Probabilistic simulation of electricity price scenarios using conditional generative adversarial networks},
	author={Walter, Viktor and Wagner, Andreas},
	journal={Energy and AI},
	volume={18},
	pages={100422},
	year={2024},
	publisher={Elsevier}
}

@article{nabila2025data,
	title={Data efficiency assessment of generative adversarial networks in energy applications},
	author={Nabila, Umme Mahbuba and Lin, Linyu and Zhao, Xingang and Gurecky, William L and Ramuhalli, Pradeep and Radaideh, Majdi I},
	journal={Energy and AI},
	volume={20},
	pages={100501},
	year={2025},
	publisher={Elsevier}
}

@article{menon2022generative,
  title={A generative approach to materials discovery, design, and optimization},
  author={, Dhruv and Ranganathan, Raghavan},
  journal={ACS omega},
  volume={7},
  number={30},
  pages={25958--25973},
  year={2022},
  publisher={ACS Publications}
}

@article{bastek2023inverse,
  title={Inverse design of nonlinear mechanical metamaterials via video denoising diffusion models},
  author={Bastek, Jan-Hendrik and Kochmann, Dennis M},
  journal={Nature Machine Intelligence},
  volume={5},
  number={12},
  pages={1466--1475},
  year={2023},
  publisher={Nature Publishing Group UK London}
}

@article{park2024inverse,
  title={Inverse design of porous materials: a diffusion model approach},
  author={Park, Junkil and Gill, Aseem Partap Singh and Moosavi, Seyed Mohamad and Kim, Jihan},
  journal={Journal of Materials Chemistry A},
  volume={12},
  number={11},
  pages={6507--6514},
  year={2024},
  publisher={Royal Society of Chemistry}
}

@article{barra2025inverse,
  title={Inverse mapping of properties to composition through generative modeling for designing molten salts},
  author={Barra, Julian and Chahal, Rajni and Banerjee, Shubhojit and Lupo Pasini, Massimiliano and Irle, Stephan and Lam, Stephen},
  journal={npj Computational Materials},
  volume={11},
  number={1},
  pages={190},
  year={2025},
  publisher={Nature Publishing Group UK London}
}

@article{jiang2025diffusion,
  title={A diffusion-model-based approach for forecasting energy demand in New Zealand’s transport sector},
  author={Jiang, Shang and Tran, Cong Quoc and Keyvan-Ekbatani, Mehdi},
  journal={Applied Energy},
  volume={400},
  pages={126617},
  year={2025},
  publisher={Elsevier}
}

@article{zhang2024generating,
  title={Generating synthetic net load data with physics-informed diffusion model},
  author={Zhang, Shaorong and Cheng, Yuanbin and Yu, Nanpeng},
  journal={arXiv preprint arXiv:2406.01913},
  year={2024}
}

@inproceedings{sohl2015deep,
	title={Deep unsupervised learning using nonequilibrium thermodynamics},
	author={Sohl-Dickstein, Jascha and Weiss, Eric and Maheswaranathan, Niru and Ganguli, Surya},
	booktitle={International conference on machine learning},
	pages={2256--2265},
	year={2015},
	organization={PMLR}
}

@article{ho2020denoising,
	title={Denoising diffusion probabilistic models},
	author={Ho, Jonathan and Jain, Ajay and Abbeel, Pieter},
	journal={Advances in neural information processing systems},
	volume={33},
	pages={6840--6851},
	year={2020}
}

@article{dhariwal2021diffusion,
	title={Diffusion models beat gans on image synthesis},
	author={Dhariwal, Prafulla and Nichol, Alexander},
	journal={Advances in neural information processing systems},
	volume={34},
	pages={8780--8794},
	year={2021}
}

@inproceedings{maze2023diffusion,
	title={Diffusion models beat gans on topology optimization},
	author={Maz{\'e}, Fran{\c{c}}ois and Ahmed, Faez},
	booktitle={Proceedings of the AAAI conference on artificial intelligence},
	volume={37},
	number={8},
	pages={9108--9116},
	year={2023}
}

@inproceedings{stypulkowski2024diffused,
	title={Diffused heads: Diffusion models beat gans on talking-face generation},
	author={Stypu{\l}kowski, Micha{\l} and Vougioukas, Konstantinos and He, Sen and Zi{\k{e}}ba, Maciej and Petridis, Stavros and Pantic, Maja},
	booktitle={Proceedings of the IEEE/CVF Winter Conference on Applications of Computer Vision},
	pages={5091--5100},
	year={2024}
}

@article{mukhopadhyay2023diffusion,
	title={Diffusion models beat gans on image classification},
	author={Mukhopadhyay, Soumik and Gwilliam, Matthew and Agarwal, Vatsal and Padmanabhan, Namitha and Swaminathan, Archana and Hegde, Srinidhi and Zhou, Tianyi and Shrivastava, Abhinav},
	journal={arXiv preprint arXiv:2307.08702},
	year={2023}
}

@article{croitoru2023diffusion,
	title={Diffusion models in vision: A survey},
	author={Croitoru, Florinel-Alin and Hondru, Vlad and Ionescu, Radu Tudor and Shah, Mubarak},
	journal={IEEE Transactions on Pattern Analysis and Machine Intelligence},
	volume={45},
	number={9},
	pages={10850--10869},
	year={2023},
	publisher={IEEE}
}

@article{ramesh2022hierarchical,
	title={Hierarchical text-conditional image generation with clip latents},
	author={Ramesh, Aditya and Dhariwal, Prafulla and Nichol, Alex and Chu, Casey and Chen, Mark},
	journal={arXiv preprint arXiv:2204.06125},
	volume={1},
	number={2},
	pages={3},
	year={2022}
}

@inproceedings{rombach2022high,
	title={High-resolution image synthesis with latent diffusion models},
	author={Rombach, Robin and Blattmann, Andreas and Lorenz, Dominik and Esser, Patrick and Ommer, Bj{\"o}rn},
	booktitle={Proceedings of the IEEE/CVF conference on computer vision and pattern recognition},
	pages={10684--10695},
	year={2022}
}

@article{saharia2022photorealistic,
	title={Photorealistic text-to-image diffusion models with deep language understanding},
	author={Saharia, Chitwan and Chan, William and Saxena, Saurabh and Li, Lala and Whang, Jay and Denton, Emily L and Ghasemipour, Kamyar and Gontijo Lopes, Raphael and Karagol Ayan, Burcu and Salimans, Tim and others},
	journal={Advances in neural information processing systems},
	volume={35},
	pages={36479--36494},
	year={2022}
}

@inproceedings{wyatt2022anoddpm,
	title={Anoddpm: Anomaly detection with denoising diffusion probabilistic models using simplex noise},
	author={Wyatt, Julian and Leach, Adam and Schmon, Sebastian M and Willcocks, Chris G},
	booktitle={Proceedings of the IEEE/CVF Conference on Computer Vision and Pattern Recognition},
	pages={650--656},
	year={2022}
}

@article{tashiro2021csdi,
	title={Csdi: Conditional score-based diffusion models for probabilistic time series imputation},
	author={Tashiro, Yusuke and Song, Jiaming and Song, Yang and Ermon, Stefano},
	journal={Advances in Neural Information Processing Systems},
	volume={34},
	pages={24804--24816},
	year={2021}
}

@article{jiang2024fast,
	title={Fast Denoising Diffusion Probabilistic Models for Medical Image-to-Image Generation},
	author={Jiang, Hongxu and Imran, Muhammad and Ma, Linhai and Zhang, Teng and Zhou, Yuyin and Liang, Muxuan and Gong, Kuang and Shao, Wei},
	journal={arXiv preprint arXiv:2405.14802},
	year={2024}
}

@misc{zhou2024super,
	title={Super resolution tau PET using a diffusion probabilistic model},
	author={Zhou, Ziyuan and Song, Tzu-An and Yang, Fan and Lois, Cristina and Johnson, Keith and Dutta, Joyita},
	year={2024},
	publisher={Soc Nuclear Med}
}

@article{zhang2024interpretable,
	title={An Interpretable Latent Denoising Diffusion Probabilistic Model for Fault Diagnosis Under Limited Data},
	author={Zhang, Tian and Lin, Jing and Jiao, Jinyang and Zhang, Han and Li, Hao},
	journal={IEEE Transactions on Industrial Informatics},
	year={2024},
	publisher={IEEE}
}

@article{devlin2024diffusion,
	title={Diffusion model approach to simulating electron-proton scattering events},
	author={Devlin, Peter and Qiu, Jian-Wei and Ringer, Felix and Sato, Nobuo},
	journal={Physical Review D},
	volume={110},
	number={1},
	pages={016030},
	year={2024},
	publisher={APS}
}

@article{li2024sensing,
  title={Sensing anomaly of photovoltaic systems with sequential conditional variational autoencoder},
  author={Li, Ding and Zhang, Yufei and Yang, Zheng and Jin, Yaohui and Xu, Yanyan},
  journal={Applied Energy},
  volume={353},
  pages={122124},
  year={2024},
  publisher={Elsevier}
}

@article{bregere2020simulating,
  title={Simulating tariff impact in electrical energy consumption profiles with conditional variational autoencoders},
  author={Br{\'e}g{\`e}re, Margaux and Bessa, Ricardo J},
  journal={IEEE Access},
  volume={8},
  pages={131949--131966},
  year={2020},
  publisher={IEEE}
}

@article{zheng2023conditional,
  title={Conditional variational autoencoder informed probabilistic wind power curve modeling},
  author={Zheng, Zhong and Yang, Luoxiao and Zhang, Zijun},
  journal={IEEE Transactions on Sustainable Energy},
  volume={14},
  number={4},
  pages={2445--2460},
  year={2023},
  publisher={IEEE}
}

@article{park2025multi,
	title={Multi-modal conditional diffusion model using signed distance functions for metal-organic frameworks generation},
	author={Park, Junkil and Lee, Youhan and Kim, Jihan},
	journal={Nature Communications},
	volume={16},
	number={1},
	pages={34},
	year={2025},
	publisher={Nature Publishing Group UK London}
}

@article{dong2023short,
  title={Short-term wind power scenario generation based on conditional latent diffusion models},
  author={Dong, Xiaochong and Mao, Zhihang and Sun, Yingyun and Xu, Xinzhi},
  journal={IEEE Transactions on Sustainable Energy},
  volume={15},
  number={2},
  pages={1074--1085},
  year={2023},
  publisher={IEEE}
}

@article{fu2024creating,
  title={Creating synthetic energy meter data using conditional diffusion and building metadata},
  author={Fu, Chun and Kazmi, Hussain and Quintana, Matias and Miller, Clayton},
  journal={Energy and Buildings},
  volume={312},
  pages={114216},
  year={2024},
  publisher={Elsevier}
}

@article{wang2024customized,
  title={Customized load profiles synthesis for electricity customers based on conditional diffusion models},
  author={Wang, Zhenyi and Zhang, Hongcai},
  journal={IEEE Transactions on Smart Grid},
  volume={15},
  number={4},
  pages={4259--4270},
  year={2024},
  publisher={IEEE}
}

\end{document}